%% file: main.tex
\documentclass[10pt,twocolumn,letterpaper]{article}

\usepackage{algorithm}
\usepackage{algpseudocode} 
\usepackage{pifont}

\usepackage[final]{cvpr}

\definecolor{cvprblue}{rgb}{0.21,0.49,0.74}
\usepackage[pagebackref,breaklinks,colorlinks,allcolors=cvprblue]{hyperref}

\def\paperID{42785}
\def\confName{CVPR}
\def\confYear{2026}

\newcommand\blfootnote[1]{%
  \begingroup
  \renewcommand\thefootnote{}\footnote{#1}%
  \addtocounter{footnote}{-1}%
  \endgroup
}

\title{FoleyDesigner: Immersive Stereo Foley Generation with Precise Spatio-Temporal Alignment for Film Clips}

\author{
    Mengtian Li$^{1,2}$  \quad
    Kunyan Dai$^{1}$ \quad
    Yi Ding$^{1}$ \quad
    Ruobing Ni$^{1}$ \quad
    Ying Zhang$^{1}$ \quad
    \\[0.5em]
    Wenwu Wang$^{3\dagger}$ \quad
    Zhifeng Xie$^{1,2\dagger}$
    \\[0.5em]
    $^{1}$Shanghai Film Academy, Shanghai University \\
    $^{2}$Shanghai Engineering Research Center of Motion Picture Special Effects \\
    $^{3}$University of Surrey, UK \\[-0.5 em]
}

\begin{document}
\maketitle
\blfootnote{$^\dagger$Corresponding authors.}

\input{sec/0_abstract}    
\input{sec/1_intro}
\input{sec/2_relative}

\input{sec/3_method}
\input{sec/4_exp}

\input{sec/5_conclusion}
\section{Acknowledgments}
This work was supported by the National Natural Science Foundation of China (Grant No. 62402306), the NaturalScience Foundation of Shanghai (Grant No. 24ZR1422400, Grant No. 25ZR1401130), the Open Research Project ofthe State Key Laboratory of Industrial Control Technol-ogy, China (Grant No. ICT2024B72).

{
    \small
    \bibliographystyle{ieeenat_fullname}
    \bibliography{main}
}

\input{sec/X_suppl}
\end{document}

%% file: sec/0_abstract.tex
\begin{abstract}
Foley art plays a pivotal role in enhancing immersive auditory experiences in film, yet manual creation of spatio-temporally aligned audio remains labor-intensive. We propose \textbf{FoleyDesigner}, a novel framework inspired by professional Foley workflows, integrating film clip analysis, spatio-temporally controllable Foley generation, and professional audio mixing capabilities.
FoleyDesigner employs a multi-agent architecture for precise spatio-temporal analysis. It achieves spatio-temporal alignment through latent diffusion models trained on spatio-temporal cues extracted from video frames, combined with large language model (LLM)-driven hybrid mechanisms that emulate post-production practices in film industry. 
To address the lack of high-quality stereo audio datasets in film, we introduce \textbf{FilmStereo}, the first professional stereo audio dataset containing spatial metadata, precise timestamps, and semantic annotations for eight common Foley categories. 
For applications, the framework supports interactive user control while maintaining seamless integration with professional pipelines, including 5.1-channel Dolby Atmos systems compliant with ITU-R BS.775 standards, thereby offering extensive creative flexibility.
Extensive experiments demonstrate that our method achieves superior spatio-temporal alignment compared to existing baselines, with seamless compatibility with professional film production standards. The project page is available at \url{https://gekiii996.github.io/FoleyDesigner/}.

\end{abstract}

%% file: sec/1_intro.tex
\section{Introduction}

\label{sec:intro}

\begin{figure}[t]
\centering
\includegraphics[width=1\columnwidth]{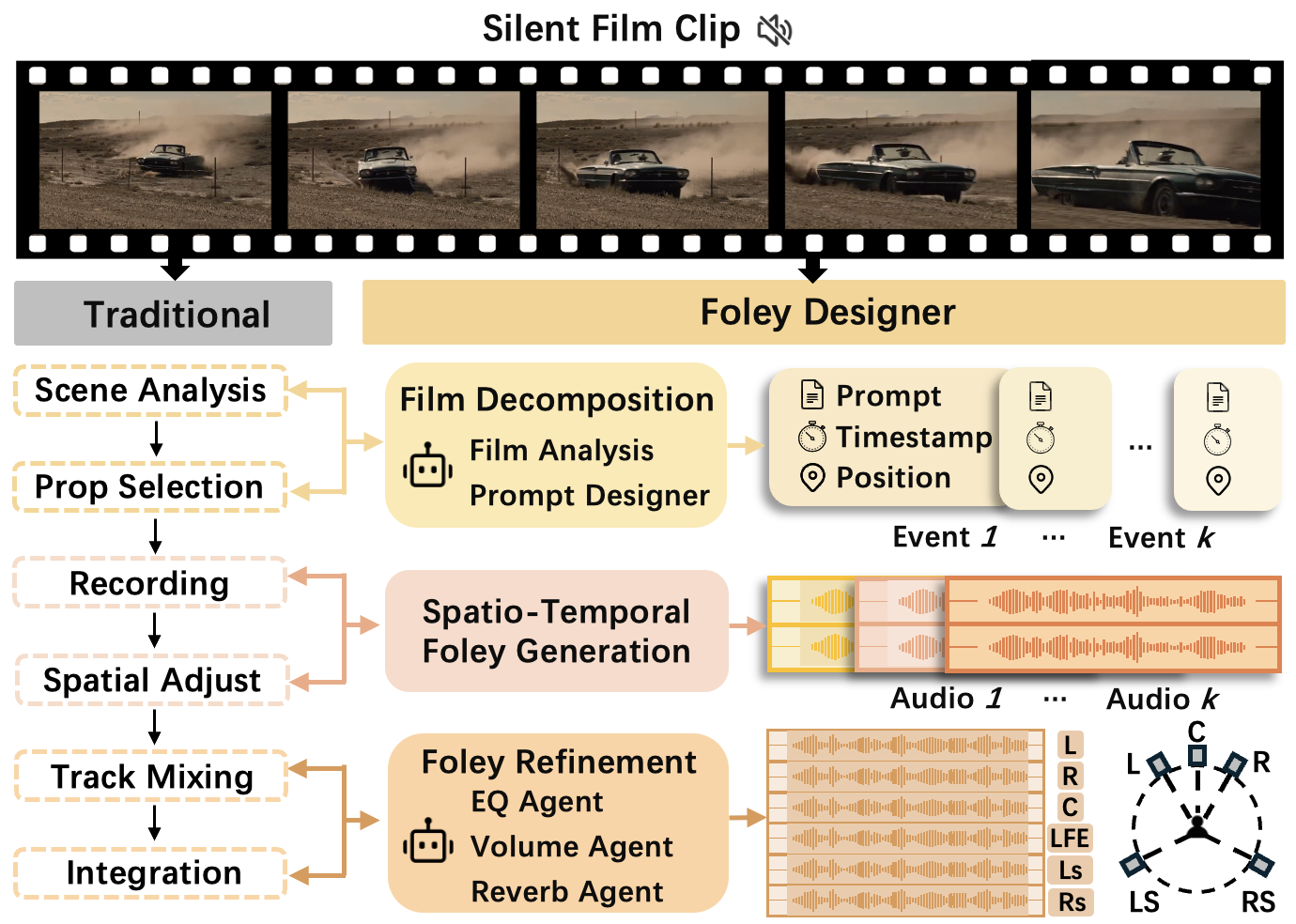} 
\caption{\textbf{FoleyDesigner Overview.} The left column detailing the actual steps of a human Foley designer. The right column presents the corresponding simulated functional modules of FoleyDesigner, showcasing outputs at each phase, resulting in a soundtrack suitable for film use.
 }
\label{fig:teaser}
\vspace{-2mm}
\end{figure}

Stereo Foley refers to the art of creating and recording sound effects with spatial information, where sounds are positioned across the left and right audio channels to create a sense of directionality and space. In film production, the stereo Foley serves critical narrative and immersive functions. Film Foley artists must synchronize sounds with on-screen actions with frame-level precision while tracking the spatial movement of visual elements. The spatial positioning of sounds is designed according to the script and directorial vision to guide the audience's attention, convey emotions, and enhance dramatic moments. This requires careful control over timing, spatial placement, and sonic qualities to maintain audience immersion and support storytelling. However, general stereo Foley generation methods cannot meet these film production requirements.

Despite the importance of stereo Foley in film production, current audio generation methods fall short of meeting professional requirements. Existing approaches can be categorized into three groups, each with critical limitations. First, monaural generation methods such as AudioLDM2~\cite{liu2024audioldm2learningholistic}, Tango 2~\cite{majumder2024tango}, and Make-an-Audio 2~\cite{huang2023makeanaudio2temporalenhancedtexttoaudio} produce high-quality sound effects from text prompts using latent diffusion models, but they lack spatial dimensions entirely, rendering them unsuitable for stereo Foley. Second, stereo generation methods like Stable Audio~\cite{evans2024stableaudioopen}, which performs diffusion on waveform representations, can produce stereo audio but lacks precise spatial localization control. Similarly, SpatialSonic~\cite{sun2024both} and See2Sound~\cite{dagli2024see} generate stereo audio from images or text but cannot achieve frame-level temporal alignment with visual content. Third, monaural-to-stereo conversion methods such as Sep-Stereo~\cite{zhou2020sepstereovisuallyguidedstereophonic} and Mono to Binaural~\cite{parida2022beyond} depend on pre-existing monaural sources, limiting their flexibility and requiring additional production steps.

Generating film-quality Foley audio requires addressing three technical challenges: (1) \textit{Densely overlapping sound events.} Film scenes contain multiple simultaneous spectrally-temporally overlapping sound sources. Single-pass generative models cannot disentangle these complex acoustic scenes, resulting in incomplete or muddled audio outputs that fail to capture the layered nature of professional Foley. (2) \textit{Precise spatio-temporal grounding.} Current conditioning mechanisms lack explicit grounding in visual spatial cues and temporal dynamics. Text-based approaches provide only coarse directional descriptions that cannot specify continuous spatial trajectories or frame-accurate timing. Image-based methods lack temporal information and cannot capture the dynamic movement of sound sources. (3) \textit{Professional acoustic quality.} Generated audio exhibits acoustic inconsistencies that fail to meet film production standards. Mismatched reverberation between sound events, spectral masking in overlapping frequencies, and loudness imbalances cause important information to be buried in the mixed sounds, degrading cinematic immersion. These limitations prevent current audio generation systems from being used in professional film Foley workflows.


To address these challenges, we present \textbf{FoleyDesigner}, a novel framework that integrates professional Foley production pipelines to generate film-quality stereo audio from silent film clips. FoleyDesigner begins with fine-grained Foley decomposition, where Tree-of-Thought reasoning with multi-agent verification analyzes visual content and script context to produce Foley scripts that decompose complex scenes into layered sound events. These decomposed events are then processed through spatio-temporal Foley generation, where we introduce a novel spatio-temporal injection mechanism that conditions a Diffusion Transformer on sound event trajectories extracted from visual tracking, achieving frame-accurate spatio-temporal alignment with visual motion. Finally, Foley refinement and professional mixing employs a multi-agent framework where specialized diagnostic agents identify acoustic inconsistencies through complementary analysis tools and determine mixing parameters, applying professional audio engineering knowledge to ensure reverberation coherence, spectral clarity, and balanced dynamics before upmixing to 5.1 channel surround formats for film applications, as shown in Figure~\ref{fig:teaser}.

To support FoleyDesigner, we construct \textbf{FilmStereo}, the first spatial audio dataset specifically designed for film Foley generation. Unlike existing audio-visual datasets that lack spatial metadata or contain only coarse temporal alignment, FilmStereo provides stereo recordings with precise temporal annotations and spatial positioning information across eight common Foley categories. FilmStereo enables data-driven training of spatially grounded audio generation models and establishes a benchmark for evaluating film-quality Foley synthesis.

Our contributions are summarized as follows:
\begin{itemize}
\item We present \textbf{FoleyDesigner}, the first framework that integrates professional Foley production workflows through decomposition, generation, and refinement stages, and introduce \textbf{FilmStereo}, a large-scale stereo audio dataset with precise temporal and spatial annotations across eight common Foley categories.

\item We propose a spatio-temporal injection mechanism that conditions diffusion transformers on visual tracking trajectories for precise alignment, and introduce a multi-agent framework with Tree-of-Thought reasoning for Foley script validation and automated audio production, promoting automated film Foley production.

\item We demonstrate practical application into current film production workflows, where FoleyDesigner generates high-quality multi-channel (e.g. 5.1) surround soundtracks with professional film standards, improving production efficiency while maintaining professional quality. 

\end{itemize}

%% file: sec/2_relative.tex
\section{Related work}
\label{sec:Related work}

\subsection{Monaural Audio Generation}
Monaural audio generation has seen significant progress across \textbf{text-to-audio (T2A)}, \textbf{image-to-audio (I2A)}, and \textbf{video-to-audio (V2A)} tasks. In T2A, autoregressive models like AudioGen~\cite{kreuk2022audiogen} treat audio synthesis as conditional language modeling, while diffusion-based models such as AudioLDM~\cite{liu2023audioldm}, Tango2~\cite{majumder2024tango}, and Make-an-Audio2~\cite{huang2023makeanaudio2temporalenhancedtexttoaudio} leverage latent diffusion to produce audio from text prompts. For I2A, methods like ImageHear~\cite{sheffer2023hear}, CLIPSonic~\cite{dong2023clipsonic}, and V2A-Mapper~\cite{wang2024v2a} utilize CLIP features to generate audio from static images. V2A approaches, including Frieren~\cite{wang2024frieren} and MMAudio~\cite{cheng2024taming}, enhance semantic and temporal coherence with video inputs. Recent foley generation work includes Spotlighting~\cite{huang2025spotlightingpartiallyvisiblecinematic}, MultiFoley~\cite{chen2025videoguidedfoleysoundgeneration}, VideoFoley~\cite{lee2024video}, CondFoleyGen~\cite{du2023conditional}, FoleyCrafter~\cite{zhang2024foleycrafterbringsilentvideos}, and DiffFoley~\cite{luo2023diff}. However, these methods face limitations for film production. T2A and I2A approaches lack precise audio-visual synchronization required for frame-accurate foley. V2A methods remain limited to monaural output and struggle with spatial positioning. Foley generation systems produce single-channel audio often misaligned with spatial information in the visual scene. 

Our method addresses these limitations by integrating spatial cues directly into the generation process, enabling stereo foley synthesis with precise audio-visual synchronization tailored for film production workflows.

\subsection{Spatial Audio Generation}
Spatial audio generation has evolved from early stereo techniques~\cite{may2010probabilistic} to modern deep learning approaches. \textbf{Stereo audio localization} research, such as Yang and Zheng~\cite{yang2022deepear} and Cao et al.~\cite{cao2021improved}, focuses on analyzing interaural time and level differences for accurate sound positioning in single- and multi-source scenarios, but primarily addresses analysis and separation rather than controllable generation. \textbf{Mono-to-stereo conversion} methods conditioned on visual inputs~\cite{garg2021geometry, liu2024visually} and depth cues~\cite{parida2022beyond} convert existing mono audio to stereo representations, but rely on pre-existing audio sources, fundamentally limiting creative flexibility. Generative models like MusicGen~\cite{copet2023simple} and Stable Audio~\cite{evans2024stableaudioopen} generate stereo audio natively but without explicit spatial control mechanisms. Recent advances including SpatialSonic~\cite{sun2024both}, See2Sound~\cite{dagli2024see}, and OmniAudio~\cite{liu2025omniaudiogeneratingspatialaudio} improve spatial precision through text, image, or video conditioning. However, these methods neglect frame-level temporal alignment essential for synchronizing audio with visual events in film applications, and their architectures lack integration pathways required for professional post-production workflows. In contrast, FoleyDesigner provides an end-to-end pipeline that unifies stereo generation, explicit spatial control, and 5.1-channel simulation, designed specifically to meet the demands of film post-production workflows.

%% file: sec/3_method.tex
\section{Method}

\begin{figure*}[ht]
    \centering
    \includegraphics[width=\linewidth]{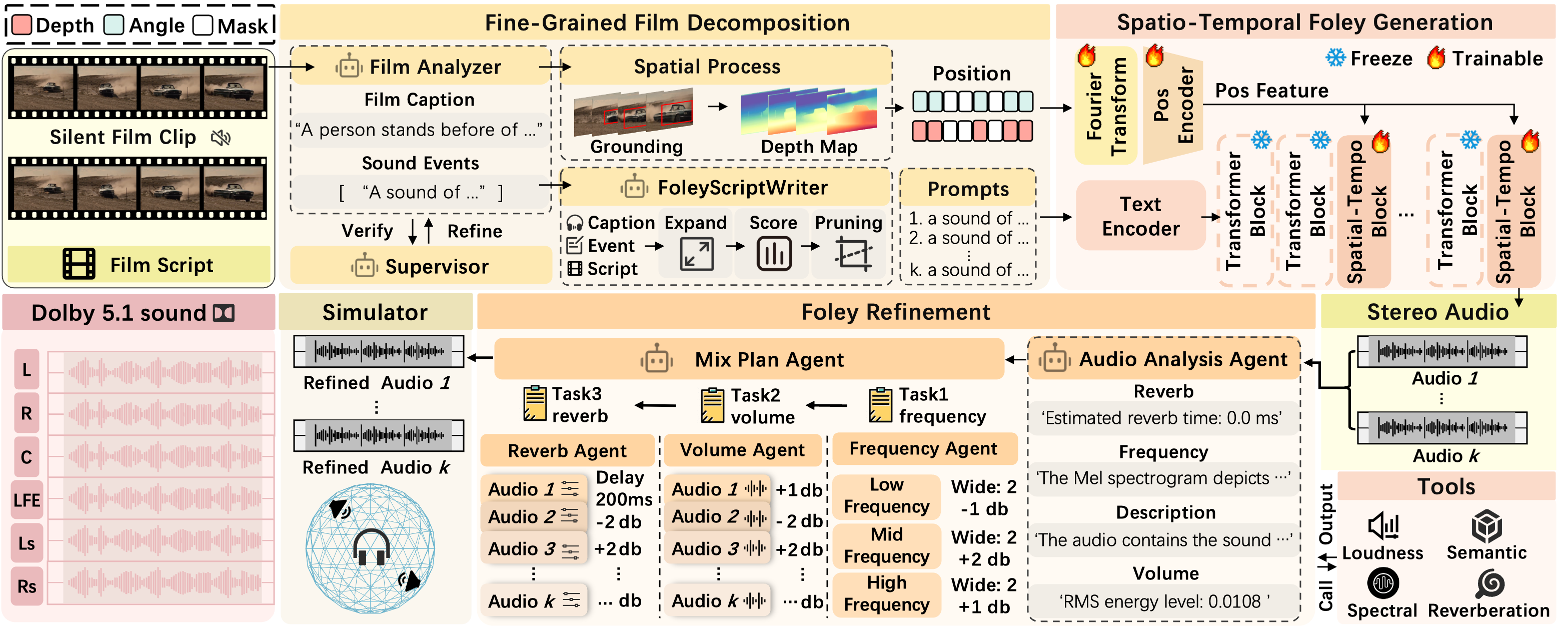}
    \caption{\textbf{FoleyDesigner Architecture.} Our pipeline for automated Foley generation consists of three stages, (1) \textit{Fine-Grained Film Decomposition}: analyzes silent video and generates hierarchical Foley scripts; (2) \textit{Spatio-Temporal Foley Generation}: produces spatially-controlled stereo audio using DiT-based diffusion conditioned on visual cues; (3) \textit{Foley Refinement}: applies multi-agent processing to refine audio quality and generate 5.1 surround output.}
    \label{fig:method}
\vspace{-4mm}
\end{figure*}

Unlike prior work that generates audio holistically without explicit scene decomposition or spatial control, we address three critical challenges: (1) densely overlapping sound events, (2) lack of audio-visual grounding control, and (3) acoustic inconsistencies in raw audio. FoleyDesigner integrates professional Foley workflows through three sequential stages, as illustrated in Figure~\ref{fig:method}. We first employ Tree-of-Thought reasoning to decompose silent film clips into verified Foley scripts specifying multiple temporally-layered sound events, enabling generation of densely overlapping soundscapes. Each event is then synthesized as stereo audio through a Diffusion Transformer conditioned on textual descriptions and spatial cues extracted from visual tracking, providing explicit grounding in visual dynamics. Finally, professional audio processing including reverberation, equalization, and dynamics control refines acoustic quality and upmixes stereo tracks to multi-channel surround formats suitable for cinematic production.

\subsection{Fine-Grained Film Decomposition for Foleys}
\textbf{Challenges.}
(1) Single-pass generative models cannot generate sound events simultaneously in densely overlapping scenarios, resulting in incomplete audio outputs. (2) Effective Foley design requires capturing physically observable events and inferring narrative elements that depend on script context.

To address these challenges, we decompose Foley generation into two agent-orchestrated modules that construct a verified Foley script through progressive refinement.

\textbf{FilmScribe} converts silent video \(\mathcal{V}\) into structured Foley script \(\mathcal{T}\) containing visual descriptions and sound event specifications. A generator agent produces initial script \(\mathcal{T}^{(0)}\) from \(\mathcal{V}\), while a validator agent verifies accuracy and completeness. The system iteratively refines through:
\begin{equation}
\mathcal{T}^{(k+1)} = \text{Generator}(\mathcal{V}, \text{Feedback}(\mathcal{T}^{(k)}, \mathcal{V})),
\end{equation}
where \(k\) denotes the iteration step, until \(\text{Validator}(\mathcal{T}, \mathcal{V}) \rightarrow \text{True}\). This closed-loop verification ensures visual-audio alignment and sound event completeness.

\textbf{FoleyScriptWriter} integrates film script \(\mathcal{F}\) and structured text \(\mathcal{T}\) to produce hierarchical Foley script \(\mathcal{S} = \{(e_i, l_i)\}\), where \(e_i\) denotes the \(i\)-th sound event and \(l_i \in \{\text{fg}, \text{bg}\}\) specifies its layer assignment. By decomposing densely overlapping events individually, each sound event can be generated sequentially, avoiding incomplete or muddled outputs. While visual analysis captures physically observable events, film script \(\mathcal{F}\) enables inference of narrative elements not directly observable from visual alone.
We employ Tree-of-Thought (ToT)~\cite{yao2023treethoughtsdeliberateproblem} reasoning over directed search graph \(\mathcal{G} = (\mathcal{N}, \mathcal{E})\), where \(\mathcal{N}\) denotes candidate script nodes and \(\mathcal{E}\) represents refinement edges (Algorithm~\ref{alg:tot}).

(1) \textit{Expand.} Starting from root node encoding $(\mathcal{V}, \mathcal{F})$, an expansion agent generates child nodes representing candidate scripts grounded in Foley design principles: diegetic source separation, semantic alignment with visuals, and emotional modulation matching film tone.

(2) \textit{Score.} Each candidate receives the score from $\text{Score}(\mathcal{S}, \mathcal{V}, \mathcal{F})$ which evaluates visual-audio correspondence, foreground-background separation, and film tone consistency. Misalignments, ambiguous layering, or emotional conflicts reduce respective scores.

(3) \textit{Optimization.} Search terminates immediately when $\text{Score}(\mathcal{S}, \mathcal{V}, \mathcal{F}) > \tau$. Otherwise, the refinement is performed as follows:
\begin{itemize}
    \item \textit{Refinement}: For correctable issues, such as overlapping layers or ambiguous event descriptions, we spawn sub-nodes by applying targeted adjustments.
    \item \textit{Regeneration}: For fundamental failures, such as emotional misalignment or structural incoherence, we create sibling nodes with revised constraints.
\end{itemize}

Pruning retains top-$k$ candidates per level. Termination occurs when $\text{Score}(\mathcal{S}, \mathcal{V}, \mathcal{F}) > \tau$, depth exceeds $d_{\max}$, or the branch budget is exhausted. This strategy delivers a Foley script ensuring physical fidelity, perceptual clarity, and narrative coherence for downstream diffusion synthesis.

\begin{algorithm}[t]
\small
\caption{Tree-of-Thought Foley Script Generation}
\label{alg:tot}
\begin{algorithmic}[1]
\Require Video \(\mathcal{V}\), structured text \(\mathcal{T}\), film script \(\mathcal{F}\), branching factor \(k\), beam size \(b\), depth limit \(d_{\max}\), threshold \(\tau\)
\Ensure Optimal Foley script \(\mathcal{S}^*\)
\State Initialize search graph \(\mathcal{G} = (\mathcal{N}, \mathcal{E})\) with root \(n_0 = (\mathcal{V}, \mathcal{T}, \mathcal{F})\)
\State Initialize candidate set \(\mathcal{C} \gets \{n_0\}\), depth \(d \gets 0\)
\While{\(d < d_{\max}\) \textbf{and} \(\max_{n \in \mathcal{C}} \text{Score}(\mathcal{S}_n, \mathcal{V}, \mathcal{F}) < \tau\)}
    \State \textbf{Expand}: For each \(n \in \mathcal{C}\), generate \(k\) children via refinement or regeneration
    \State \textbf{Score}: Evaluate \(\text{Score}(\mathcal{S}_n, \mathcal{V}, \mathcal{F}) = w_1 s_{\text{align}} + w_2 s_{\text{layer}} + w_3 s_{\text{emotion}}\)
    \State \textbf{Prune}: Retain top-\(b\) nodes by score \(\rightarrow \mathcal{C}\)
    \State \(d \gets d + 1\)
\EndWhile
\State \Return \(\mathcal{S}^* = \arg\max_{n \in \mathcal{C}} \text{Score}(\mathcal{S}_n, \mathcal{V}, \mathcal{F})\)
\end{algorithmic}
\end{algorithm}

\subsection{Spatio-Temporal Foley Generation}\label{sec:stereo-audio-generation}

\textbf{Motivation.} Text-conditioned models face limitations for Foley generation: (1) they cannot specify precise spatial cue for sound trajectories, (2) lack grounding in video frame geometry, and (3) cannot ensure frame-accurate temporal alignment. We address this by extracting spatio-temporal cues from video frames and conditioning diffusion generation through spatial and temporal controls.

\subsubsection{Spatio-Temporal Cue Extraction}

To enable spatial positioning of sound events, we extract depth and azimuth information from \(N\) keyframes \(\mathcal{K} = \{I_1, I_2, \ldots, I_N\}\) sampled from the input video \(\mathcal{V}\). We employ a vision language model (VLM)~\cite{bai2025qwen25vltechnicalreport} to localize sound sources by annotating bounding boxes \(\mathcal{B} = \{b_1, b_2, \ldots, b_N\}\), where each \(b_i\) corresponds to the spatial extent of a sound event in keyframe \(I_i\).

For depth estimation, we leverage~\cite{yang2024depth} to generate depth maps \(\mathbf{D}_i \in \mathbb{R}^{H \times W}\) for each keyframe, where \(H\) and \(W\) denote the frame height and width. For each bounding box \(b_i\), we compute the average depth within the box to obtain a scalar depth value \(d_i \in \mathbb{R}\). The azimuth angle \(\theta_i\) is derived from the horizontal center \(x_i \in [0, W]\) of the bounding box:
\begin{equation}
\label{eq:azimuth}
\theta_i = \arctan\left(\frac{x_i - W/2}{d_i}\right) \cdot \frac{180^\circ}{\pi} + 90^\circ,
\end{equation}
This formulation maps relative horizontal positions into azimuth angles, with \(\theta_i \in [0^\circ, 180^\circ]\) representing the sound source's angular position.

To ensure temporal synchronization, we detect sound event timestamps~\cite{xie2024sonicvisionlm} to generate a binary activation vector \(\mathbf{c} = \{c_1, c_2, \ldots, c_T\} \in \{0,1\}^T\), where \(T\) is the total number of video frames and \(c_t \in \{0, 1\}\) indicates whether a sound event occurs at frame \(t\). The spatial position sequence \(\mathcal{X} = \{\mathbf{x}_i = (d_i, \theta_i)\}_{i=1}^N\) extracted from keyframes is temporally interpolated to match the video frame rate, yielding \(\{\mathbf{x}_t\}_{t=1}^T\), and masked by the activation vector:
\begin{equation}
\label{eq:spatiotemporal_cue}
\mathbf{p}_t = c_t \cdot \mathbf{x}_t, \quad \mathcal{P} = \{\mathbf{p}_t\}_{t=1}^T,
\end{equation}
where \(\mathbf{p}_t \in \mathbb{R}^2\) encodes the depth and azimuth of the active sound event at frame \(t\).

\subsubsection{DiT-Based Conditional Generation}

We build upon Stable Audio Open~\cite{evans2024stableaudioopen}, a DiT-based latent diffusion model, conditioning it on text prompt \(\mathbf{c}_{\text{text}}\) and spatio-temporal cues \(\mathcal{P}\). We introduce a \textbf{position-aware injection mechanism} via cross-attention to achieve precise spatio-temporal alignment.

\textbf{Positional Feature Encoding.}
To enhance the expressiveness of raw spatio-temporal cues, we apply Fourier feature transformation~\cite{tancik2020fourier} to each position vector:
\begin{equation}\label{eq:fourier}
\gamma(\mathbf{p}_t) = \left[ \cos(2\pi \mathbf{B} \mathbf{p}_t); \sin(2\pi \mathbf{B} \mathbf{p}_t) \right] \in \mathbb{R}^{2m},
\end{equation}
where \(\mathbf{B} \in \mathbb{R}^{m \times 2}\) is a random projection matrix sampled from \(\mathcal{N}(0, \sigma^2)\), \(m\) is the number of frequency bands, and \([\cdot;\cdot]\) denotes concatenation.

To incorporate temporal activation information, we modulate the Fourier features with the binary mask \(c_t\):
\begin{equation}
\label{eq:mask_modulation}
\tilde{\gamma}(\mathbf{p}_t) = c_t \cdot \gamma(\mathbf{p}_t) + \epsilon \cdot \gamma(\mathbf{p}_t),
\end{equation}
where \(\epsilon = 0.1\) is a small constant that preserves weak positional information during inactive frames.

We design a convolutional encoder to process the modulated features, matching the temporal compression ratio \(r\) of the audio latent space to ensure positional embeddings maintain temporal alignment with audio latents. Each downsampling block consists of 1D convolution, group normalization, and SiLU activation. The final positional embeddings are obtained as:
\begin{equation}
\label{eq:posembedding}
\mathbf{E}_{\text{pos}} = \text{PosEncoder}(\{\tilde{\gamma}(\mathbf{p}_t)\}_{t=1}^T) \in \mathbb{R}^{T' \times d_{\text{emb}}},
\end{equation}
where \(T' = T / r\) is the compressed temporal dimension and \(d_{\text{emb}}\) is the positional embedding dimension.

\textbf{Injection Blocks.}
Our diffusion backbone processes noisy latent representations \(\mathbf{z}_{\ell} \in \mathbb{R}^{T' \times d_{\text{latent}}}\) through transformer blocks, where \(d_{\text{latent}}\) denotes the latent feature dimension of the diffusion model, distinct from \(d_{\text{emb}}\).
To inject spatio-temporal control, we insert an \textbf{injection block} after every four standard DiT blocks at layers \(\ell \in \{3, 7, 11, 15, 19, 23\}\). Each injection block performs cross-attention between latent features and layer-normalized positional embeddings:
\begin{equation}
\label{eq:injection}
\mathbf{z}'_{\ell} = \text{InjBlock}(\mathbf{z}_{\ell}, \text{LN}(\mathbf{E}_{\text{pos}})),
\end{equation}
where \(\text{InjBlock}(\cdot)\) is a transformer block with cross-attention. This design allows the model to integrate spatial awareness across different depths of the network.

\subsection{Foley Refinement and Professional Mixing}

\textbf{Challenges.}
(1) \textit{Acoustic inconsistency}: mismatched reverberation or frequency characteristics break film immersion; (2) \textit{Spectral masking}: overlapping frequencies muddy sonic clarity; (3) \textit{Loudness imbalance}: improper volume levels cause important Foley to be buried.

To address these challenges, we introduce a multi-agent post-processing framework that emulates the collaborative framework of professional Foley teams.

\textbf{Foley Analysis.} To diagnose acoustic issues beyond semantics, the agent combines perceptual and quantitative tools. For each generated Foley track $a_i$, the agent extracts composite feature representation $\mathbf{f}_i = [\mathbf{f}_{sem}, \mathbf{f}_{spec}, f_{rev}, f_{loud}]$, where $\mathbf{f}_{sem}$ denotes semantic embeddings from an Audio-LLM~\cite{chu2023qwen}, $\mathbf{f}_{spec}$ represents spectral patterns from Mel-spectrograms analyzed through a VLM~\cite{bai2025qwen25vltechnicalreport}, $f_{rev}$ is the computed reverberation time, and $f_{loud}$ is the measured integrated loudness. This combines semantics with objective measurements for comprehensive acoustic diagnosis.

\textbf{Mixing Planner.} The planner agent performs track-wise diagnosis through cross-modal validation, inter-track balance analysis, and quality assessment. It produces a mixing plan \(\boldsymbol{\Pi} = \{(i, \mathcal{O}_i)\}_{i=1}^{N_{\text{track}}}\), where \(\mathcal{O}_i \subseteq \{\texttt{reverb}, \texttt{eq}, \texttt{dyn}\}\) specifies required operations for track \(i\).

\textbf{Specialist Execution.} The plan is dispatched to three specialist agents that operate collaboratively to determine concrete processing parameters. The \textit{Reverberation Specialist} analyzes spatial relationships and determines reverberation parameters matched to scene properties. The \textit{Equalization Specialist} examines spectral overlap to determine frequency band adjustments, minimizing spectral masking and maintaining sonic clarity. The \textit{Dynamics Specialist} evaluates relative loudness levels and determines gain adjustments to prevent important Foley elements from being buried. Together, these specialists ensure that individual adjustments contribute to overall acoustic coherence.

\textbf{Multi-Channel Surround Upmixing.} To meet film Foley standards, we extend stereo output to 5.1 surround format \(\{\text{FL}, \text{FR}, \text{C}, \text{SL}, \text{SR}, \text{LFE}\}\) following ITU-R BS.775 configuration~\cite{series2010multichannel}. We adopt a channel-wise upmixing strategy that preserves spatial dynamics without simulating room acoustics. Stereo channels \(\mathbf{s}_L\) and \(\mathbf{s}_R\) are directly mapped to front left (FL) and front right (FR) channels. Center (C), surround left (SL), and surround right (SR) channels are derived from weighted mixes of the stereo signal, with distinct coefficients applied to simulate spatial positions while maintaining energy balance. The low-frequency effects (LFE) channel is produced by low-pass filtering the full mix below 120\,Hz:
\begin{equation}
\mathbf{s}_{\text{LFE}}(t) = \text{LPF}(\mathbf{s}_{\text{mix}}(t),\ 120\,\text{Hz}),
\end{equation}
where \(\mathbf{s}_{\text{mix}}(t) = \mathbf{s}_L(t) + \mathbf{s}_R(t)\). This produces 5.1-channel output that maintains spatial and temporal accuracy of the stereo source while meeting professional film standards.

%% file: sec/4_exp.tex
\section{Dataset: FilmStereo}
\begin{figure}[h]
    \centering
    \includegraphics[width=\columnwidth]{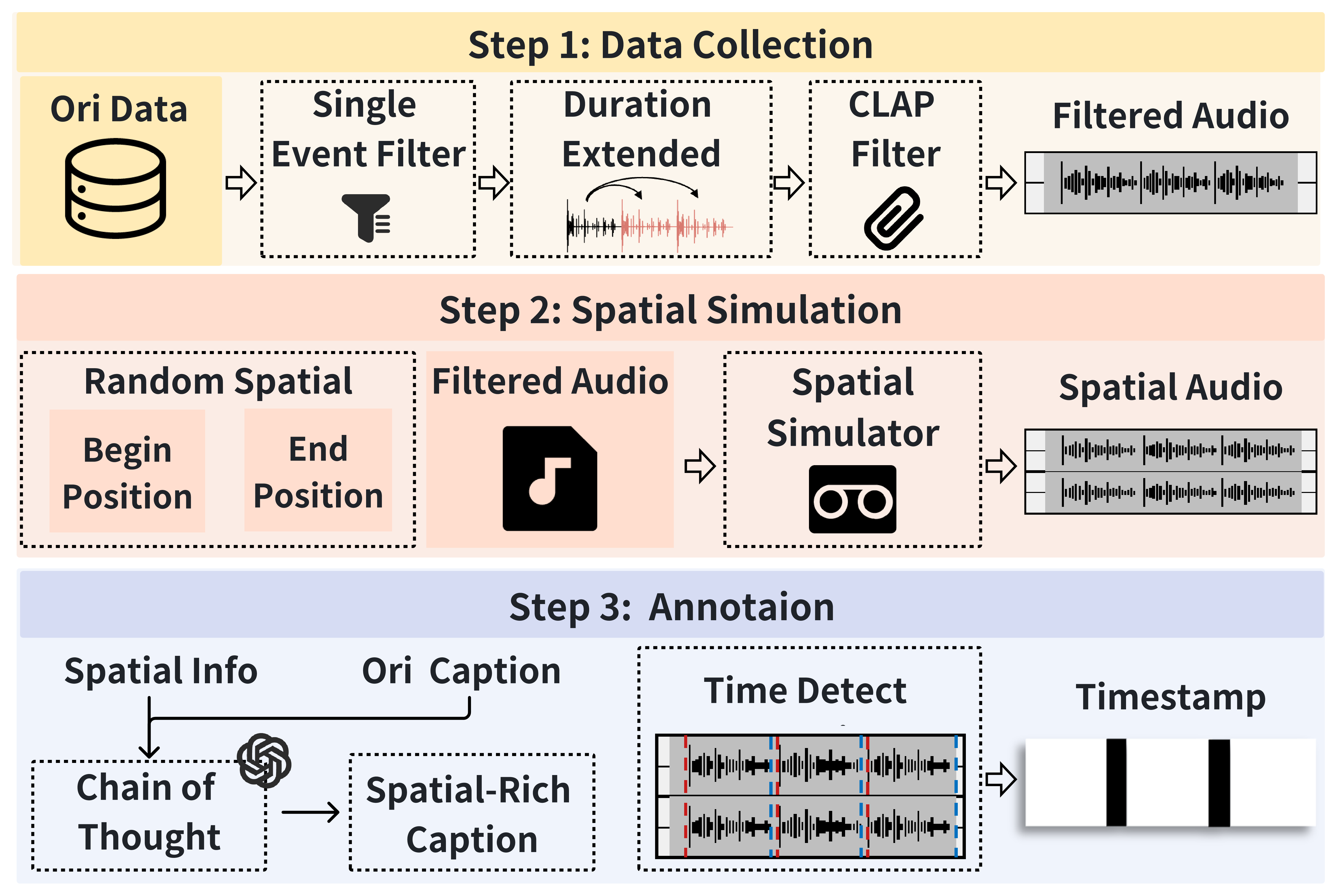}
    \caption{\textbf{FilmStereo Construction.} Our four-step pipeline for creating spatially and temporally annotated audio data. (1) Audio filtering and extension, (2) Spatial simulation with random positioning, (3) Spatial-rich caption generation via chain-of-thought, and (4) Temporal annotation with event detection.}
    \label{fig:data_pipline}
    \vspace{-2mm}
\end{figure}

To enable controllable spatial audio generation for film production, we construct \textbf{FilmStereo}, a stereo audio dataset integrating spatial captions, temporal annotations, and stereo audio. Existing datasets lack this combination, providing temporal annotations without spatial information or spatial audio without temporal alignment. Figure \ref{fig:data_pipline} illustrates our construction pipeline.

\noindent\textbf{Collection.}
We collect audio from public repositories across 8 categories (23 subcategories). The preprocessing includes: (1) filtering multi-event samples, (2) spectral denoising (-40 dB), (3) loop-padding to 8--10s, and (4) CLAP-based verification ($\tau = 0.35$). The resulting FilmStereo dataset contains 166 hours across 14,784 samples.

\noindent\textbf{Spatial Simulation.} We model azimuth using five frontal regions ($\pm15^\circ$, $\pm45^\circ$, $0^\circ$) aligned with human localization acuity and depth using three zones (near-field: 0--2m, mid-field: 2--5m, far-field: $>$5m) based on psychoacoustic principles. Using gpuRIR~\cite{Diaz_Guerra_2020} with 16--18 cm interaural distance, we generate room impulse responses for static and dynamic sources, with balanced distributions across object sizes, motion types, and spatial positions. Environmental reverberation is applied using presets.

\noindent\textbf{Annotation.} Spatial captions are generated by GPT-4 with chain-of-thought prompting, integrating sound descriptions with spatial parameters (azimuth, depth, reverberation). For temporal alignment, we detect amplitude peaks in denoised audio to identify event onsets, applying adaptive thresholds based on signal-to-noise ratio. Interval thresholds determine whether segments qualify as distinct sound events, generating start and end timestamps to ensure frame-accurate synchronization with visual content during film post-production workflows.

\section{Experiments}

\textbf{Configuration.}
The complete training process of the framework is divided into two stages: (1) stereo mel-spectrogram VAE training, and (2) DiT-based diffusion model training with spatio-temporal control injection. All stages use FilmStereo datasets with learning rate $3\times10^{-5}$, batch size 8, on NVIDIA A6000 GPUs.

\noindent\textbf{Metrics.}
We evaluate our framework in the following three aspects:
(1) \textit{Audio Quality}: Inception Score (IS)~\cite{salimans2016improvedtechniquestraininggans}, KL Divergence~\cite{heusel2018ganstrainedtimescaleupdate}, Fréchet Audio Distance (FAD)~\cite{kilgour2019frechetaudiodistancemetric}, and CLAP score~\cite{wu2024largescalecontrastivelanguageaudiopretraining} assess the perceptual quality and semantic alignment of generated audio;
(2) \textit{Spatio-Temporal Alignment}: GCC-MAE and CRW-MAE~\cite{sun2024both} measure spatial localization accuracy, Fréchet Stereo Audio Distance (FSAD)~\cite{sun2024both} evaluates stereo quality, and Intersection over Union (IoU) quantifies temporal precision;
(3) \textit{Cinematic Foley Quality}: ImageBind Score~\cite{girdhar2023imagebindembeddingspacebind} measures audio-visual semantic coherence. AV-Sync~\cite{synchformer2024iashin} evaluates synchronization accuracy. For professional film foley evaluation, we introduce \textbf{Sonic Richness Score (SRS)} and \textbf{Cinematic Clarity Score (CCS)}. We use audio-capable MLLMs~\cite{chu2023qwen} to evaluate sonic layering diversity and perceptual separation quality in professional film contexts.

\subsection{Quantitative Results}

\textbf{Audio Quality.} To evaluate the generated audio quality, we convert stereo audio to monaural signals by averaging the two channels, then assess the quality using standard metrics.
Our method achieves the highest CLAP score of \textbf{0.679} and lowest FAD of \textbf{1.88}, outperforming SpatialSonic by \textbf{1.0\%} and \textbf{2.6\%}, and Stable Audio~\cite{evans2024stableaudioopen} by \textbf{14.3\%} and \textbf{20.7\%}. This demonstrates semantic alignment and perceptual quality for Foley sound design.
While our IS score is lower than SpatialSonic, IS measures sample diversity rather than quality or semantic accuracy. Our focus on spatial coherence and text-audio alignment produces contextually appropriate outputs. Results are summarized in Table~\ref{Quantitative}.

\begin{figure*}[th]
    \centering
    \includegraphics[width=\linewidth]{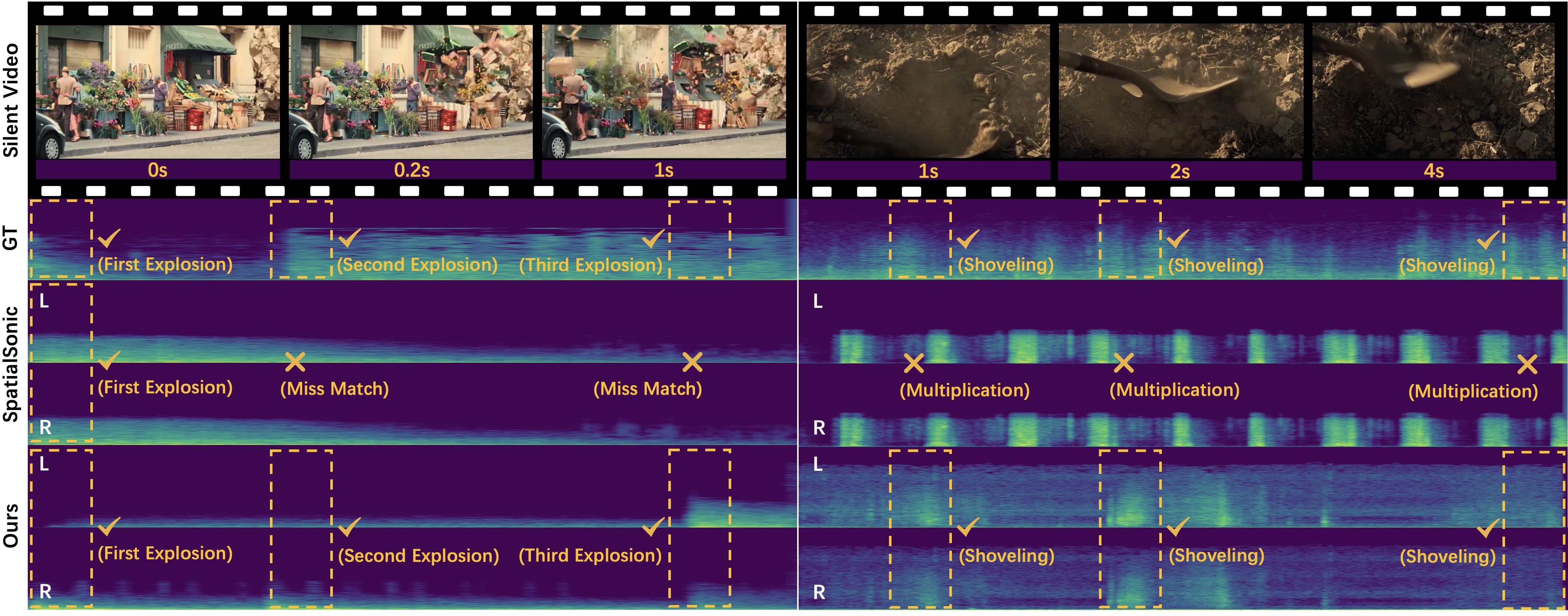}
    \caption{\textbf{Qualitative Results.} Qualitative comparison showing temporal alignment between video events and generated audio across two scenarios. Left: explosion sequence with three distinct events. Right: excavation scene with repetitive shoveling actions. Checkmarks indicate successful synchronization, while crosses mark temporal misalignment.}
    \label{fig:case_temporal}
    \vspace{-4mm}
\end{figure*}

\begin{table}[h]
\small
\begin{center}
\caption{Audio Quality. Metrics include Inception Score, KL Divergence, Fréchet Audio Distance, and CLAP score for audio quality assessment. \textbf{Best} and \underline{second-best} results are highlighted. $\downarrow$ indicates lower is better, $\uparrow$ indicates higher is better.}
\begin{tabular}{lccccccc}
\hline
\textbf{Method} & \textbf{IS $\uparrow$} & \textbf{KL $\downarrow$} & \textbf{FAD $\downarrow$} & \textbf{CLAP $\uparrow$} \\
\hline
Stable Audio\hfill~\cite{evans2024stableaudioopen}      & 10.50          & 1.86          & 2.37          & 0.594          \\
SpatialSonic\hfill~\cite{sun2024both}        & \textbf{13.79} & \underline{1.37} & \underline{1.93} & \underline{0.672} \\
\hline
Ours                                         & \underline{12.36} & \textbf{1.40} & \textbf{1.88} & \textbf{0.679} \\
\hline
\end{tabular}
\label{Quantitative}
\vspace{-4mm}
\end{center}
\end{table}

\begin{table}[h]
\small
\begin{center}
\caption{Spatio-Temporal Alignment Results. Metrics include GCC and CRW for spatial accuracy, FSAD for stereo quality, and IoU for temporal alignment. \textbf{Best} and \underline{second-best} results are highlighted. $\downarrow$ indicates lower is better, $\uparrow$ indicates higher is better.}
\begin{tabular}{lcccc}
  \hline
  \textbf{Method} & \textbf{GCC $\downarrow$} & \textbf{CRW $\downarrow$} & \textbf{FSAD $\downarrow$} & \textbf{IoU $\uparrow$} \\
  \hline
  Stable Audio\hfill~\cite{evans2024stableaudioopen}    & 61.17          & 51.44          & 0.343          & 24.5          \\
  See2Sound\hfill~\cite{dagli2024see}        & 60.03          & 51.17          & 0.291          & 21.3          \\
  SpatialSonic\hfill~\cite{sun2024both}      & \underline{49.20} & \underline{36.87} & \underline{0.163} & \underline{27.8} \\
  \hline
  Ours                                       & \textbf{48.79} & \textbf{34.23} & \textbf{0.138} & \textbf{32.2} \\
  \hline
\end{tabular}
\label{Spatial}
\end{center}
\vspace{-4mm}
\end{table}

\textbf{Spatio-Temporal Alignment.}
To evaluate spatial accuracy and temporal alignment, we assess stereo audio output using metrics in Table~\ref{Spatial}.
FoleyDesigner achieves the best performance across metrics, demonstrating the effectiveness of our position-aware injection mechanism.
For spatial accuracy, our method achieves the lowest GCC MAE (\textbf{48.79}) and CRW MAE (\textbf{34.23}), outperforming SpatialSonic by \textbf{0.8\%} and \textbf{7.2\%}.
The FSAD score (\textbf{0.138}) confirms high-quality stereo imaging with channel separation. For temporal alignment, FoleyDesigner achieves the highest IoU score (\textbf{32.2}), improving by \textbf{15.8\%} over SpatialSonic.
These results validate that our position-aware injection mechanism is effective in incorporating spatial positioning and temporal dynamics for Foley sound synthesis.

\begin{table}[h]
\caption{Film Foley Performance. Evaluation on film clips covering audio-visual synchronization (ImageBind Score, AV-Sync) and cinematic quality (Sonic Richness Score, Cinematic Clarity Score). \textbf{Best} and \underline{second-best} results are highlighted.}
\small
\vspace{-2mm}
\begin{center}
  \begin{tabular}{lccccc}
    \hline
    \textbf{Method} & \textbf{IB $\uparrow$} & \textbf{SRS $\uparrow$} & \textbf{CCS $\uparrow$} & \textbf{AV-Sync $\uparrow$} \\
    \hline
    Stable Audio\hfill~\cite{evans2024stableaudioopen}  & 0.216 & 5.31 & 5.8 & 0.512 \\
    See2Sound\hfill~\cite{dagli2024see}      & 0.105 & 3.03 & 3.0 & 0.601 \\
    SpatialSonic\hfill~\cite{sun2024both}    & 0.251 & 5.91 & 4.5 & 0.545 \\
    \hline
    Ours                                     & \textbf{0.402} & \textbf{8.27} & \textbf{6.2} & \textbf{0.726} \\
    \hline
  \end{tabular}
  \label{tab:temporal-results}
\end{center}
\vspace{-2mm}
\end{table}

\textbf{Film Foley Performance.} To evaluate the effectiveness of our method in real film scenarios, we assess the generated audio across multiple cinematic dimensions using film clips, as shown in Table~\ref{tab:temporal-results}.
FoleyDesigner achieves the best performance across all metrics. Our method obtains the highest ImageBind Score (\textbf{0.402}) and AV-Sync score (\textbf{0.726}), outperforming SpatialSonic by 60.2\% and 33.2\% respectively, attributed to the position-aware injection mechanism that enables explicit grounding in visual spatial cues and temporal dynamics.
The superior SRS (\textbf{8.27}) and CCS (\textbf{6.2}) scores, improving by 39.9\% and 37.8\%, validate our film-oriented design. The sonic richness stems from fine-grained film decomposition, while the cinematic clarity results from multi-agent refinement that optimizes acoustic balance across tracks through reverberation, equalization, and dynamics processing.

\subsection{Qualitative Results}
We present qualitative comparisons on two representative film scenarios, demonstrating \textbf{superior temporal synchronization} and \textbf{semantic alignment} compared to existing approaches. As shown in Figure~\ref{fig:case_temporal}, the visualization displays silent input clips, ground truth spectrograms, SpatialSonic baseline outputs, and our results.

In the explosion sequence, our method accurately captures the timing and intensity dynamics of three explosions with precise temporal correspondence to visual events. In contrast, SpatialSonic correctly aligns with the first explosion but fails to maintain temporal accuracy for subsequent events, showing temporal drift with mismatches for the second and third explosions.
Similarly, for the excavation scene, SpatialSonic exhibits ``Multiplication'' artifacts, generating repetitive sounds that do not correspond to the actual visual events, whereas our method successfully generates appropriate shoveling sounds that align with the digging actions shown in the video frames.
These qualitative results validate that our approach achieves superior performance in both synchronization accuracy and acoustic quality across different film scenarios. Comprehensive quantitative results and details are provided in Supplementary Materials.

\subsection{Ablation Study}
To validate the effectiveness of spatio-temporal cues (STC) in FoleyDesigner, we conduct an ablation study comparing the full model against a baseline without STC. The results are shown in Table~\ref{tab:ablation}. The baseline configuration without STC shows higher errors across all metrics.
Incorporating STC yields substantial improvements: GCC reduces by \textbf{21.3\%} and CRW decreases by \textbf{38.8\%}, demonstrating critical gains in spatio-temporal alignment. The FAD improves by \textbf{12.1\%}, reflecting better perceptual quality.

These results validate that spatio-temporal cues are essential for generating high-quality Foley audio. The consistent improvements across all metrics demonstrate the effectiveness of our position-aware injection mechanism design in achieving superior performance.

\begin{table}[h]
\caption{Ablation study results on the FilmStereo dataset. STC refers to spatio-temporal cues including trajectory information and spatial positioning prompts. \textbf{Best} results are highlighted.}
\small
\centering
\begin{tabular}{lcccc}
\hline
\textbf{Configuration} & \textbf{GCC $\downarrow$} & \textbf{CRW $\downarrow$} & \textbf{FSAD $\downarrow$} & \textbf{FAD $\downarrow$} \\
\hline
w/o STC    & 62.02          & 55.89          & 0.297          & 2.14          \\
Full Model & \textbf{48.79} & \textbf{34.23} & \textbf{0.138} & \textbf{1.88} \\
\hline
\end{tabular}
\label{tab:ablation}
\vspace{-2mm}
\end{table}


\subsection{Human Evaluation}
To assess the perceived quality of generated foleys, we conducted offline (5.1 surround) and online (stereo) human evaluations. Offline tests involved 12 participants evaluating the 5.1 output of FoleyDesigner, while the online study involved 53 participants using stereo playback. For baseline comparisons, we evaluated against See-2-Sound~\cite{dagli2024see}, Stable Audio~\cite{evans2024stableaudioopen}, and SpatialSonic~\cite{sun2024both}.
Participants rated each method across five dimensions: (1) \textit{immersion}; (2) \textit{emotional alignment} with scene atmosphere; (3) \textit{temporal alignment} with visual events; (4) \textit{spatial alignment} relative to visual sources; (5) \textit{timbre consistency} with film content.

\begin{figure}[th]
    \centering
    \includegraphics[width=1\columnwidth]{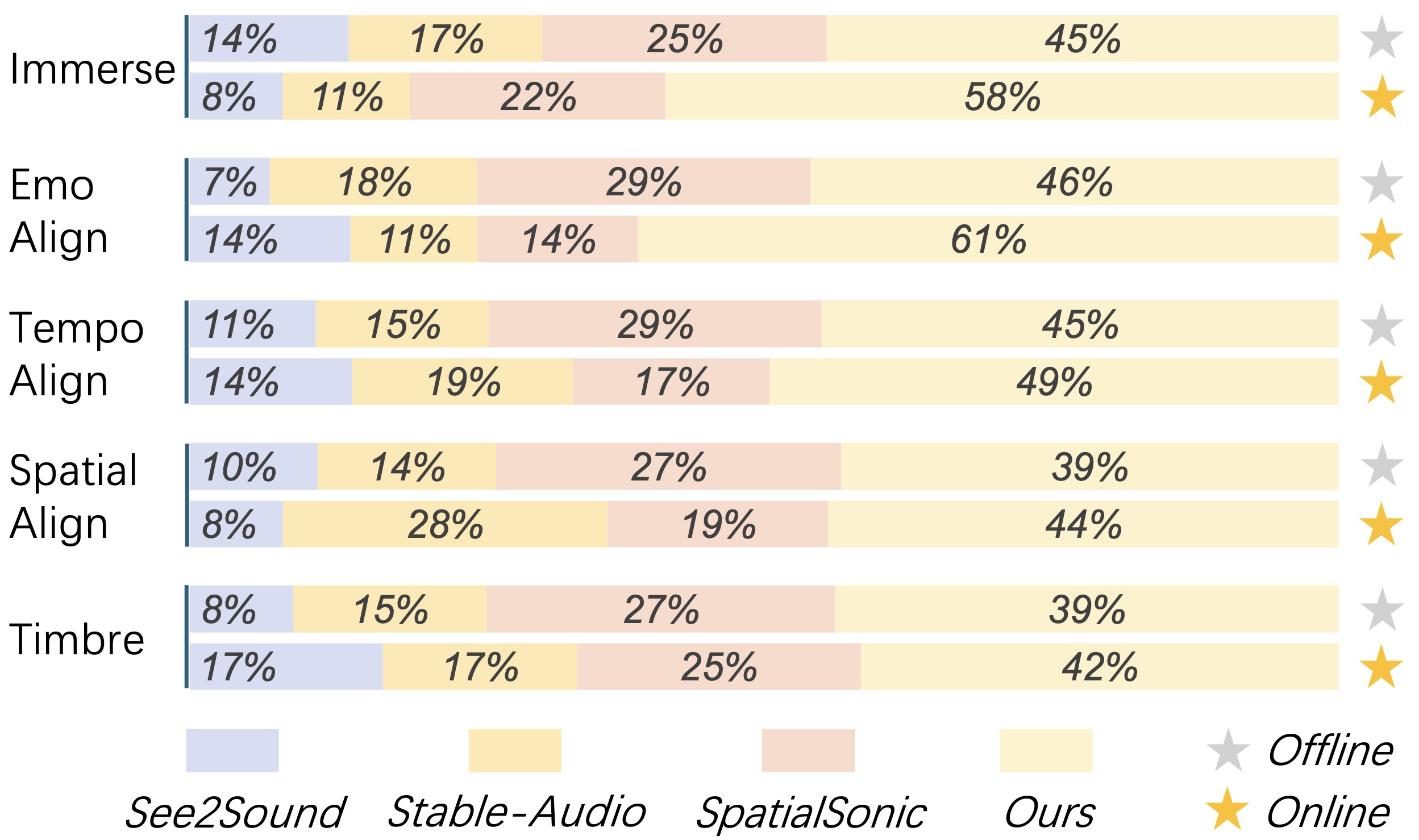}
    \caption{\textbf{Human Evaluation Results.} We compared the selection ratio of four methods from five perspectives: (1) Immerse, (2) Emo Align, (3) Tempo Align, (4) Spatial Align, and (5) Timbre.}
    \label{fig:userstudy_online}
    \vspace{-2mm}
\end{figure}

Our method demonstrates performance across all dimensions in both evaluation settings. FoleyDesigner achieves the highest preference rates in emotional alignment (61\% online preference) and immersion (58\% online preference), indicating that our spatio-temporal cues capture spatial accuracy and semantic and affective qualities of film audio. The preference distributions are presented in Figure~\ref{fig:userstudy_online}, showing advantages over existing approaches across all evaluation criteria. These results validate that our method generates spatially-aware foley audio with enhanced realism and user experience compared to baselines.

\section{Conclusion}
We have presented \textbf{FoleyDesigner}, a novel framework for generating spatio-temporally aligned stereo audio for film clips. Drawing inspiration from professional Foley workflows, our framework decomposes complex acoustic scenes into hierarchical scripts through Tree-of-Thought reasoning. It integrates multi-stage modules including film clip perception, stereo audio generation, and multi-agent mixing. Crucially, our spatio-temporal conditioning mechanism extracts depth and azimuth cues from visual tracking and injects them via position-aware cross-attention, ensuring synchronization with on-screen movements.

To support stereo Foley generation, we introduced the annotated \textbf{FilmStereo} dataset. Experimental results demonstrate that FoleyDesigner achieves superior performance in audio quality, spatial alignment, and temporal accuracy compared to existing methods. Furthermore, it demonstrates strong potential for diverse real-world applications, including film post-production and immersive sound design for virtual reality, supporting efficient and controllable Foley generation.

Despite these advances, we observe that generation performance can degrade in scenes with densely overlapping concurrent sound events (e.g., simultaneous footsteps, object interactions, and ambience), which leads to spatial localization errors. To address this limitation, our future work will focus on enhancing visual understanding through more robust multi-object tracking and hierarchical reasoning.

%% file: sec/X_suppl.tex
\clearpage
\setcounter{page}{1}
\setcounter{section}{0}
\setcounter{figure}{0}
\setcounter{table}{0}
\renewcommand{\thesection}{\arabic{section}}
\renewcommand{\thefigure}{\arabic{figure}}
\renewcommand{\thetable}{\arabic{table}}
\maketitlesupplementary
\begin{figure*}
    \centering
    \includegraphics[width=1\linewidth]{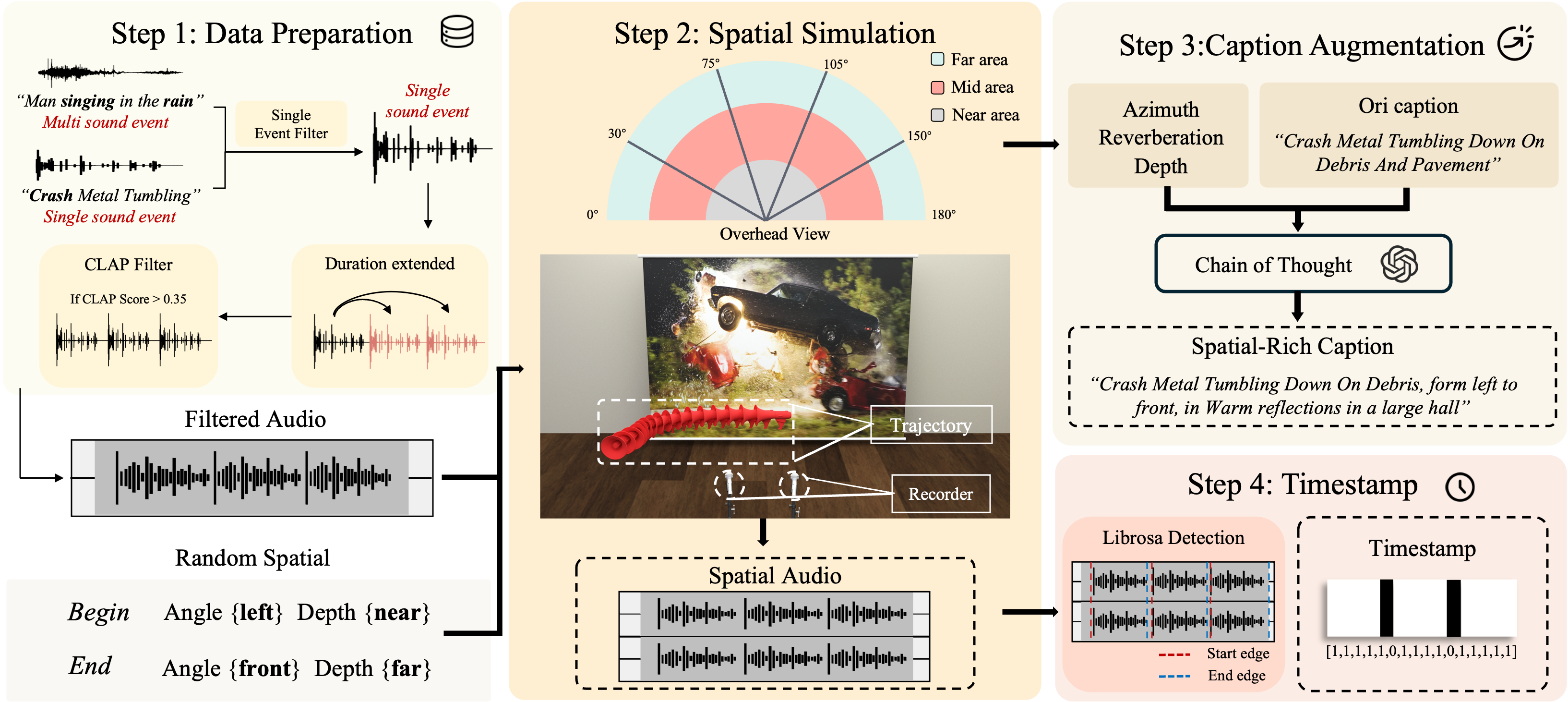}
    \label{fig:datapipeline}
    \caption{\textbf{FilmStereo Dataset Pipeline.} The process begins with sourcing data using randomly sampled parameters to define sound event attributes, followed by a simulated sound design scenario in Step 2 to generate film foley annotations. The resulting data undergoes manual verification to ensure quality and accuracy.}
\end{figure*}

\section{FilmStereo Dataset}

Current audio datasets predominantly focus on monaural sound, overlooking the pivotal role of stereophonic audio in enhancing film immersion. As summarized in Table \ref{data_table}, existing datasets typically lack stereophonic recordings or precise temporal annotations. This limitation compels sound designers to manually craft spatial effects from monaural sources, incurring substantial creative and computational overhead. To address this gap, we introduce the \textbf{FilmStereo} stereo dataset, which integrates stereo audio, spatial captions, timestamps.

\begin{table}[h]
\centering
\renewcommand{\arraystretch}{1.2}
\setlength\tabcolsep{0.8mm}
\begin{tabular}{cccc}
    \hline
    \textbf{Dataset} & \textbf{Duration (hours)} & \textbf{Temporal} & \textbf{Spatial} \\
    \hline
    LAION-Audio  & 4.3K  & \ding{55} & \ding{55} \\
    AudioCaps    & 110   & \ding{55} & \ding{55} \\
    VGG-Sound    & 550   & \ding{55} & \ding{55} \\
    AudioTime    & 15.3  & \ding{51} & \ding{55} \\
    CompA-order  & 1.5   & \ding{51} & \ding{55} \\
    Simstereo    & 116   & \ding{55} & \ding{51} \\
    FAIR-Play    & 5.2   & \ding{55} & \ding{51} \\
    MUSIC        & 23    & \ding{55} & \ding{51} \\
    \hline
    FilmStereo (ours) & 166 & \ding{51} & \ding{51} \\
    \hline
\end{tabular}
\caption{\textbf{Comparison of existing datasets.} FilmStereo provide temporal and spatial annotations simultaneously.}
\vspace{-4.5mm}
\label{data_table}
\end{table}

\subsection{Data Collection and Preparation}

To support film foley generation tasks, we organized our dataset into eight common sound categories, subdivided into 23 subcategories based on typical sound sources used in film post-production. As shown in Table~\ref{category_distribution}, the dataset covers diverse sound types ranging from human actions and environmental sounds to mechanical and impact sounds. Data were collected from publicly available repositories with samples distributed across these categories to ensure comprehensive coverage of film audio requirements.

The data preprocessing pipeline involved multiple quality control steps to ensure dataset integrity. Initially, we filtered out samples with captions indicating multiple concurrent sound events to ensure the validity of subsequent spatial simulations. The acquired audio clips were typically noisy, so we set a threshold of -40 dB to filter out silent segments and applied spectral subtraction denoising algorithms to each recording. Short-duration sounds (less than 2 seconds) were extended to 8–10 seconds through seamless loop-padding using overlap-add techniques, promoting temporal consistency. To verify semantic alignment between audio content and textual descriptions, we employed the CLAP model to compute text-audio similarity scores, discarding samples with scores below $\tau = 0.35$. This rigorous preprocessing resulted in a curated collection of high-quality audio samples suitable for spatial audio simulation.

\begin{table}[t]
\centering
\renewcommand{\arraystretch}{1.2}
\begin{tabular}{cccc}
    \hline
    \textbf{Category} & \textbf{Percentage (\%)} \\
    \hline
    Ambience \& Environments & 4.70 \\
    Bio \& Organic & 12.87 \\
    Cultural \& Abstract & 21.87 \\
    Foley \& Physical Interactions & 19.01 \\
    Designed \& Abstract & 17.95 \\
    Dynamic Systems & 6.88 \\
    Impact \& Destruction & 12.59 \\
    Mechanical \& Technology & 4.13 \\
    \hline
\end{tabular}
\caption{\textbf{Percentage distribution.} The table shows the proportion of audio clips belonging to each of the eight major sound design categories.}
\label{category_distribution}
\vspace{-4.5mm}
\end{table}

\subsection{Spatial Audio Simulation}

Guided by insights from psychoacoustic research, we modeled two fundamental spatial perception attributes: azimuth and depth. The azimuth was discretized into five frontal regions (±15°, ±45°, 0°) to correspond with the acuity of human lateral localization capabilities. Depth perception was nonlinearly categorized into three distinct zones—near-field (0–2 meters), mid-field (2–5 meters), and far-field (greater than 5 meters)—based on frequency-dependent attenuation patterns commonly observed in professional Foley practice. These derived acoustic parameters were employed to simulate stereo audio in a perceptually realistic manner that aligns with human auditory perception.

Building upon established room acoustics simulation techniques, we utilized gpuRIR for the room impulse response generation process. During this simulation, we assumed a standard interaural distance of 16–18 cm, without explicitly accounting for the head shadow effect or head-related transfer functions (HRTFs), as these factors were deemed secondary for the scope of this study. The simulation pipeline comprised two distinct stages to handle different types of sound sources. For static sound sources, we performed random sampling of azimuth and depth parameters within the defined ranges. For dynamic sources, trajectories were computed based on spatial position parameters, with intermediate positions refined using linear interpolation to ensure smooth spatial transitions. After the initial spatial simulation, we applied environmental reverberation effects using VST3 plugin presets\footnote{\url{https://www.voxengo.com/group/free-vst-plugins-download/}}, including hall, room, chamber, and plate reverbs, to match film environmental contexts and enhance the realism of the generated audio.

\subsection{Annotation}

To enhance the utility of audio datasets for film design applications, we transformed raw captions into spatially-aware descriptions enriched with comprehensive spatial and acoustic information. Sound designers typically infer sound event types, spatial positions, and reverberation effects based on the specific requirements of a film sequence, necessitating detailed spatial descriptions. We extracted parameters such as azimuth, depth, and reverberation effects directly from the audio processing operations performed during the spatial audio simulation pipeline. Using GPT-4, we employed a chain-of-thought prompting strategy to generate captions that seamlessly integrate sound event descriptions with their corresponding spatial and acoustic properties, ensuring alignment with professional sound design workflows.

To align the rhythm of sound with visual elements in film audio production, we introduced temporal annotations comprising precise start and end timestamps that define the exact timing of sound events within each audio clip. We detected amplitude peaks in the denoised audio signals to identify the onset of sound events, applying adaptive thresholds to filter out background noise and silent segments based on these detected peaks. Additionally, we established interval thresholds to determine whether a segment qualifies as a distinct sound event, thereby generating corresponding start and end timestamps with millisecond precision. This temporal annotation process ensures that the generated audio can be precisely synchronized with visual content during film post-production workflows.

\subsection{Dataset Statistics}

\begin{figure}[h]
    \centering
    \includegraphics[width=0.8\linewidth]{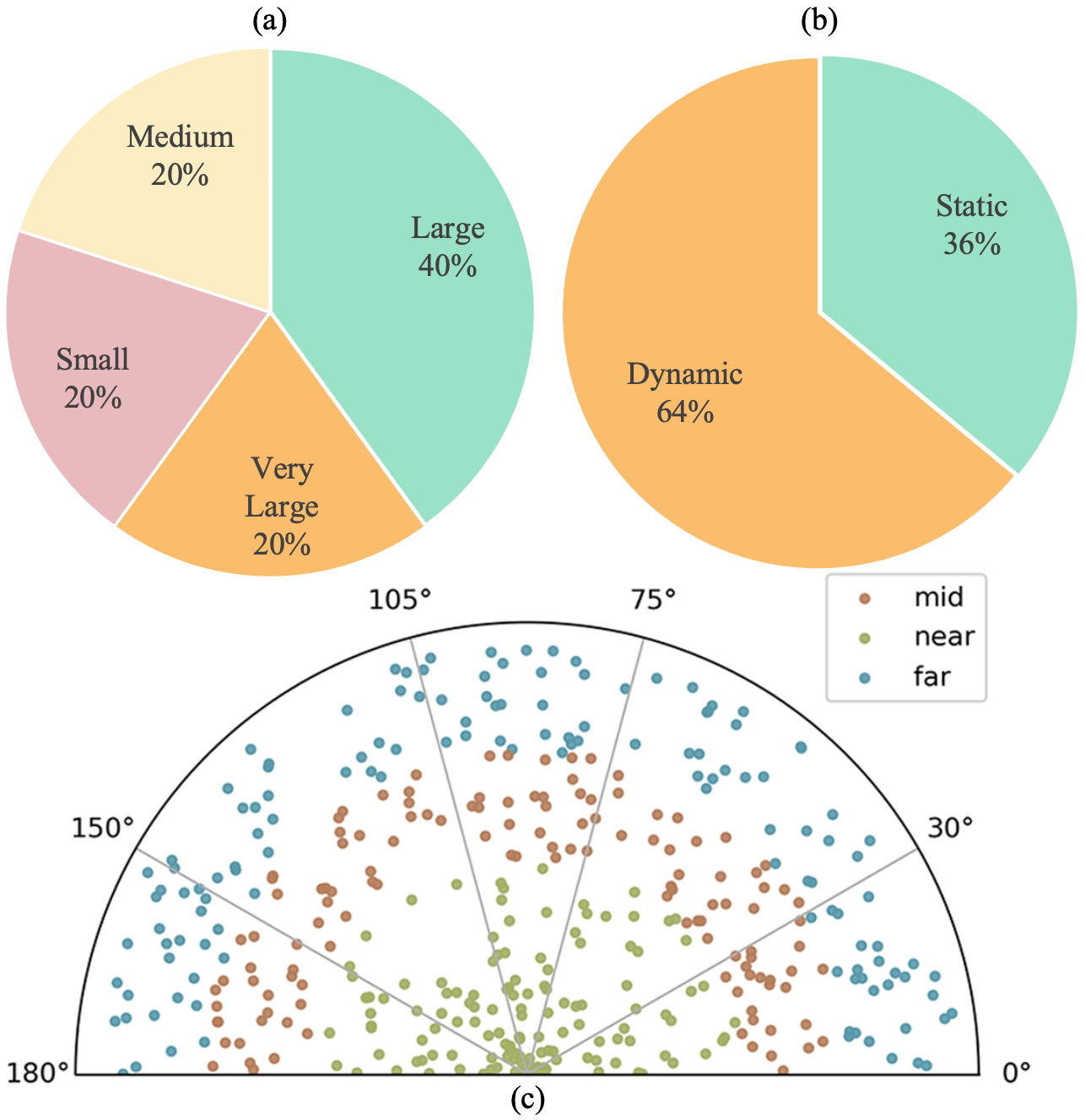}
    \caption{\textbf{FilmStereo Distribution Analysis.} (a) Room size distribution. (b) Motion type distribution. (c) Spatial positioning across azimuth angles and depth zones.}
    \label{data_distribution}
    \vspace{-4.5mm}
\end{figure}

The complete FilmStereo dataset contains 42.3 hours of stereo audio data distributed across 14,784 samples spanning the eight primary categories. As illustrated in Figure~\ref{data_distribution}, the dataset exhibits balanced distributions across multiple dimensions. In terms of object size representation, large objects constitute 40\% of the dataset, while medium, small, and very large objects each account for 20\%. The motion characteristics show a predominance of dynamic sounds (64\%) over static sounds (36\%), reflecting the dynamic nature of film audio. The spatial positioning analysis reveals a comprehensive coverage across the frontal hemisphere, with sounds distributed across near-field (green), mid-field (brown), and far-field (blue) distances, ensuring realistic spatial diversity that mirrors typical film audio scenarios.

\section{Implementation Details}

\subsection{Multi-Agent Refinement}

The complete pseudocode for our multi-agent Foley refinement pipeline is presented in Algorithm~\ref{alg:foley_refinement}. This algorithm implements the professional mixing framework described in Section 3.3 of the main paper, emulating the collaborative workflow of professional Foley teams through a four-stage process.

\textbf{Mixing Planning.} The planner agent conducts track-wise diagnosis by evaluating three critical aspects for each track. First, cross-modal validation aligns audio features with visual context to detect semantic inconsistencies between the generated sound and the visual content. Second, inter-track balance analysis examines relative loudness and spectral overlap across all tracks to identify potential masking issues where one sound obscures another. Third, acoustic quality assessment matches the reverberation characteristics and frequency distribution to the scene's spatial properties, such as indoor versus outdoor environments. Based on these diagnostic scores, the planner determines the required operations for each track, constructing a structured mixing plan that guides subsequent specialist processing.

\textbf{Specialist Execution.} The execution stage dispatches operations to three specialist agents, each focusing on a specific aspect of audio processing while maintaining awareness of inter-track relationships. The Reverberation Specialist analyzes the spatial context from visual features and determines appropriate reverberation parameters for tracks requiring spatial treatment, adjusting reverb ratio, room size, and damping values to match the acoustic environment depicted in the scene. The Equalization Specialist examines spectral overlap between tracks and determines frequency band adjustments across low, mid, and high frequency ranges to minimize masking effects and ensure each sound element occupies its appropriate spectral space. The Dynamics Specialist evaluates relative loudness levels across tracks and determines gain adjustments in decibels for balanced mixing, ensuring foreground elements maintain prominence while background sounds provide appropriate ambience without overwhelming the mix. Each specialist receives all tracks requiring its operation type simultaneously, enabling coordinated processing that ensures consistency across the entire mix rather than treating tracks in isolation.

\begin{algorithm}[t]
\small
\caption{Multi-Agent Foley Refinement and Professional Mixing}
\label{alg:foley_refinement}
\begin{algorithmic}[1]
\Require Generated Foley tracks \(\{a_i\}_{i=1}^{N}\), visual context \(\mathcal{V}\), Foley script \(\mathcal{S}\)
\Ensure Refined and mixed Foley audio \(A_{\text{final}}\)

\State \textbf{Stage 1: Foley Analysis}
\For{\(i = 1\) to \(N\)}
    \State Extract semantic embeddings: \(\mathbf{f}_{\text{sem}} \gets \text{AudioLLM}(a_i)\)
    \State Extract spectral features: \(\mathbf{f}_{\text{spec}} \gets \text{VLM}(\text{MelSpec}(a_i))\)
    \State Compute reverberation: \(f_{\text{rev}} \gets \text{RT60}(a_i)\)
    \State Measure loudness: \(f_{\text{loud}} \gets \text{LUFS}(a_i)\)
    \State Construct feature vector: \(\mathbf{f}_i \gets [\mathbf{f}_{\text{sem}}, \mathbf{f}_{\text{spec}}, f_{\text{rev}}, f_{\text{loud}}]\)
\EndFor

\State \textbf{Stage 2: Mixing Planning}
\State Initialize mixing plan: \(\Pi \gets \emptyset\)
\For{\(i = 1\) to \(N\)}
    \State Cross-modal validation: \(s_{\text{visual}} \gets \text{Align}(\mathbf{f}_i, \mathcal{V})\)
    \State Inter-track balance: \(s_{\text{balance}} \gets \text{Compare}(\mathbf{f}_i, \{\mathbf{f}_j\}_{j \neq i})\)
    \State Acoustic quality: \(s_{\text{quality}} \gets \text{Evaluate}(f_{\text{rev}}, f_{\text{loud}}, \mathcal{V})\)
    \State Determine operations: \(\mathcal{O}_i \gets \text{Diagnose}(s_{\text{visual}}, s_{\text{balance}}, s_{\text{quality}})\)
    \State Update plan: \(\Pi \gets \Pi \cup \{(i, \mathcal{O}_i)\}\)
\EndFor

\State \textbf{Stage 3: Specialist Execution}
\State Group tracks by operation type:
\State \(\mathcal{T}_{\text{reverb}} \gets \{i \mid \text{reverb} \in \mathcal{O}_i\}\)
\State \(\mathcal{T}_{\text{eq}} \gets \{i \mid \text{eq} \in \mathcal{O}_i\}\)
\State \(\mathcal{T}_{\text{dyn}} \gets \{i \mid \text{dyn} \in \mathcal{O}_i\}\)

\If{\(\mathcal{T}_{\text{reverb}} \neq \emptyset\)}
    \State \(\{\theta_{\text{rev}}^i\}_{i \in \mathcal{T}_{\text{reverb}}} \gets \text{ReverbSpecialist}(\{(a_i, \mathbf{f}_i)\}_{i \in \mathcal{T}_{\text{reverb}}}, \mathcal{V})\)
    \For{\(i \in \mathcal{T}_{\text{reverb}}\)}
        \State \(a_i \gets \text{ApplyReverb}(a_i, \theta_{\text{rev}}^i)\)
    \EndFor
\EndIf

\If{\(\mathcal{T}_{\text{eq}} \neq \emptyset\)}
    \State \(\{\theta_{\text{eq}}^i\}_{i \in \mathcal{T}_{\text{eq}}} \gets \text{EQSpecialist}(\{(a_i, \mathbf{f}_i)\}_{i \in \mathcal{T}_{\text{eq}}})\)
    \For{\(i \in \mathcal{T}_{\text{eq}}\)}
        \State \(a_i \gets \text{ApplyEQ}(a_i, \theta_{\text{eq}}^i)\)
    \EndFor
\EndIf

\If{\(\mathcal{T}_{\text{dyn}} \neq \emptyset\)}
    \State \(\{\theta_{\text{dyn}}^i\}_{i \in \mathcal{T}_{\text{dyn}}} \gets \text{DynamicsSpecialist}(\{(a_i, \mathbf{f}_i)\}_{i \in \mathcal{T}_{\text{dyn}}})\)
    \For{\(i \in \mathcal{T}_{\text{dyn}}\)}
        \State \(a_i \gets \text{ApplyDynamics}(a_i, \theta_{\text{dyn}}^i)\)
    \EndFor
\EndIf

\State \textbf{Stage 4: Final Mixing}
\State \(A_{\text{final}} \gets \text{Mix}(\{a_1, a_2, \ldots, a_N\})\)
\State \Return \(A_{\text{final}}\)
\end{algorithmic}
\end{algorithm}

\subsection{Computational Cost and Inference Time}
We provide the runtime breakdown for generating a 3-second stereo Foley clip on a single NVIDIA RTX A6000 GPU in Table~\ref{tab:runtime}. While slower than end-to-end models, our method targets professional post-production, where precision, multi-track decomposition, and human-in-the-loop controllability are prioritized over real-time generation speed.

\begin{table}[h]
\centering
\caption{Inference Time Analysis. Time consumption in seconds (s) for generating a 3s stereo clip on a single A6000 GPU.}
\small
\begin{tabular}{l|c|l}
\hline
Stage & Time (s) & Component \\
\hline
1. Visual Analysis & 2s & VLM  \\
2. Script Decomposition & 34s & LLM Agents \\
3. Audio Generation & 8s & DiT Diffusion  \\
4. Foley Refinement & 64s & LLM Agents \\
\textbf{Total} & \textbf{108s} & \textbf{vs. End-to-End ($\sim$5s)} \\
\hline
\end{tabular}
\label{tab:runtime}
\end{table}

\section{Additional Quantitative Evaluations}

To provide a more comprehensive understanding of FoleyDesigner's capabilities, we present additional quantitative experiments including extended baseline comparisons and an ablation study on our multi-agent framework.

\subsection{Extended Baseline Comparisons}
We evaluated additional state-of-the-art models, specifically Diff-Foley and FoleyCrafter. As shown in Table~\ref{tab:baselines}, our method achieves competitive generation quality compared to mono baselines. While FoleyCrafter shows strong mono performance, it lacks spatial control (indicated by GCC and CRW metrics). Furthermore, to investigate the impact of training data, we fine-tuned the mono baseline Tango on our FilmStereo dataset. Adapting its mono-native architecture proved suboptimal compared to our specialized design, yielding a higher FAD and inferior spatial alignment metrics. This confirms that simply fine-tuning mono models is insufficient for spatial Foley generation.

\begin{table}[h]
\centering
\caption{Generation Quality and Spatial Alignment. $\downarrow$ indicates lower is better, $\uparrow$ indicates higher is better.}
\small
\resizebox{0.95\linewidth}{!}{
\begin{tabular}{c|ccccc}
\hline
Method & FAD $\downarrow$ & ImgBind $\uparrow$ & IoU $\uparrow$ &GCC $\downarrow$  & CRW $\downarrow$    \\
\hline
Diff-Foley  & 1.85 & 0.383 & 31.80 & - & -  \\
FoleyCrafter & \textbf{1.69} & \textbf{0.430} & 32.00 & - & - \\
Tango (Fine-tuned) & 2.26 & 0.328 & 28.60 & 54.82 & 45.36 \\
\textbf{Ours} & 1.88 & 0.402 & \textbf{32.20} & \textbf{48.79} & \textbf{34.23} \\
\hline
\end{tabular}}
\label{tab:baselines}
\end{table}

\subsection{Multi-Agent Framework Ablation}
We performed an ablation study to validate the necessity of our multi-agent refinement framework, as shown in Table~\ref{tab:ablation}. In complex scenes, the single-stage baseline often missed background events, whereas our multi-agent framework significantly improved Event Recall. Crucially, objective acoustic metrics demonstrate the impact of our Mixing Specialists. While the RT60 Error shows a slight increase due to the inherent complexity of estimating precise reverberation from 2D visual cues in open environments, this trade-off is significantly outweighed by the substantial improvements in spectral clarity (LSD) and dynamic balance (Loudness Error) achieved by our Equalization and Dynamics Agents.

\begin{table}[h]
    \small
    \centering
    \caption{Ablation Study on Agents Framework. Best results are highlighted. 
    ER: Event Recall (\%); LSD: Log-spectral distance (dB); 
    LE: Loudness Error (LU); RT60E: RT60 Error (s).}
    \label{tab:ablation}
    \begin{tabular}{lcccc}
        \hline
        \textbf{Method} & \textbf{ER}$\uparrow$ & \textbf{LSD}$\downarrow$ & \textbf{LE}$\downarrow$ & \textbf{RT60E}$\downarrow$ \\
        \hline
        Single-Stage & 68.5\% & 34.20 & 7.63 & \textbf{0.92} \\
        Ours         & \textbf{84.2\%} & \textbf{18.42} & \textbf{2.86} & 1.08 \\
        \hline
    \end{tabular}
\end{table}

\section{User Study Details}

\subsection{Experimental Setup}

Our user study was conducted through both offline and online evaluations to comprehensively assess the perceived quality of generated foley audio.

\begin{figure}[h]
    \centering
    \includegraphics[width=1\linewidth]{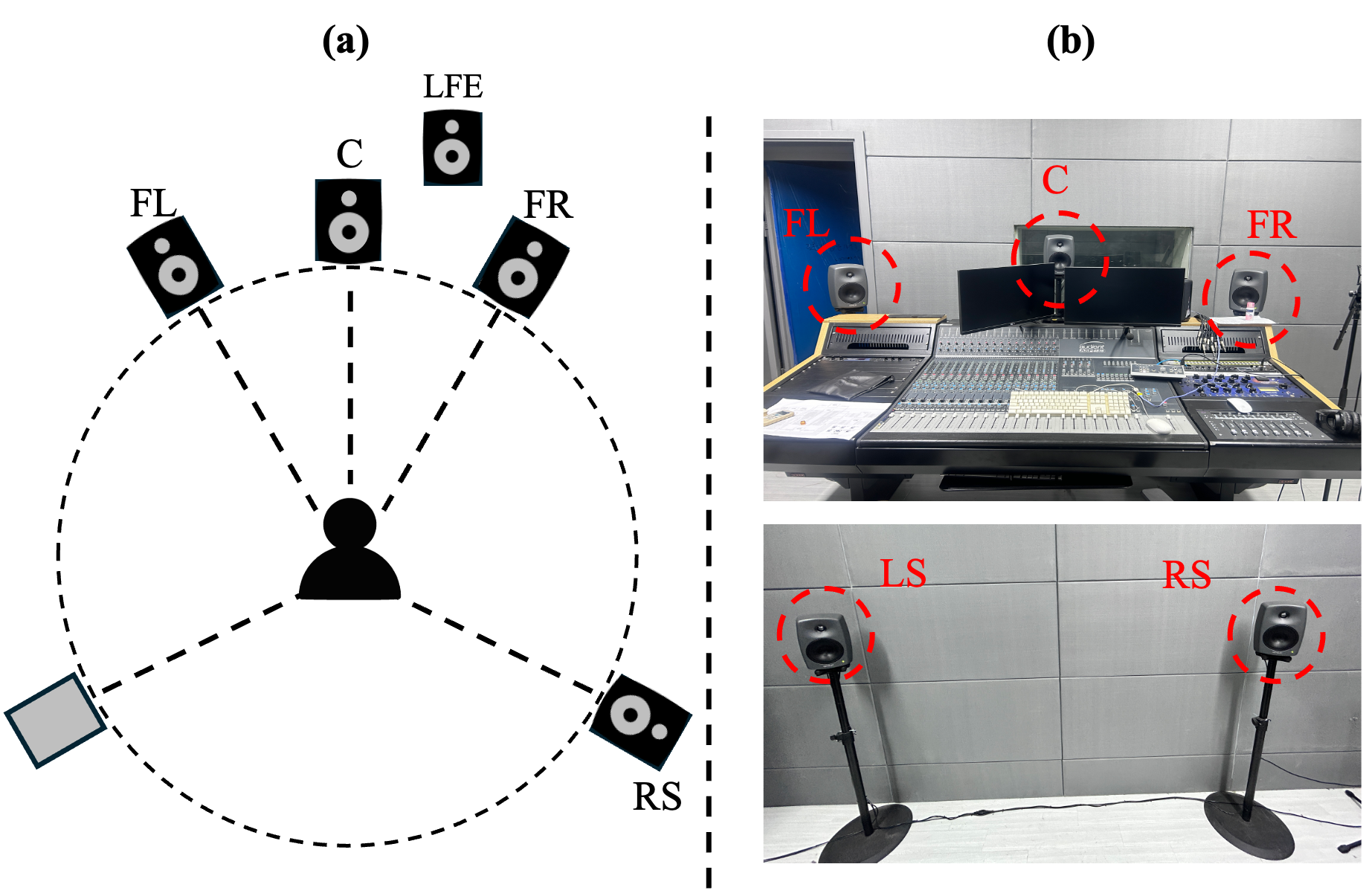}
\caption{\textbf{User study setup}. (a) Standard 5.1 surround sound speaker configuration showing Front Left (FL), Center (C), Front Right (FR), Low Frequency Effects (LFE), Left Surround (LS), and Right Surround (RS) positions. (b) Professional mixing studio environment.}
    \label{user_study_setup}
        \vspace{-4mm}
\end{figure}

\textbf{Offline Evaluation.} We recruited 12 participants with normal hearing to conduct perceptual evaluation in a professional audio mixing studio with controlled acoustic conditions. The evaluation environment featured a standard 5.1 surround sound system configured according to ITU-R BS.775-3 specifications, as illustrated in Figure~\ref{user_study_setup}.

\begin{figure}[h]
    \centering
    \includegraphics[width=0.9\linewidth]{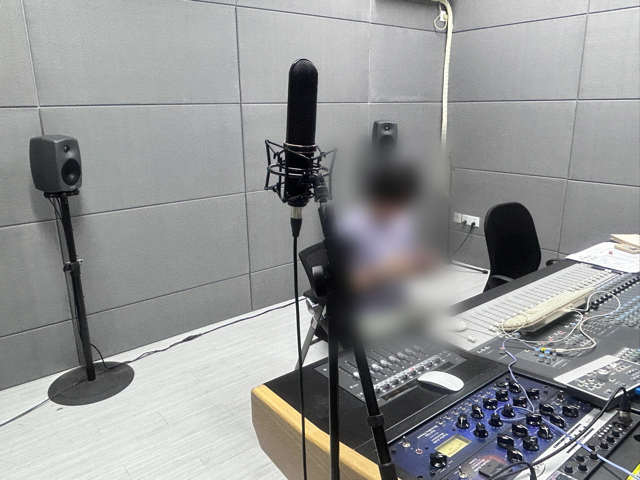}
\caption{\textbf{Participants} conducting the perceptual evaluation in the professional mixing studio environment.}
    \label{participants_setup}
    \vspace{-4mm}
\end{figure}

The listening room was acoustically optimized with controlled reverberation time and minimal background noise. Participants were positioned at the sweet spot, maintaining equal distance from all speakers. The audio playback system utilized monitors with flat frequency response to ensure accurate sound reproduction, as shown in Figure~\ref{participants_setup}.

To ensure the reliability of our subjective metrics, we strictly followed a training session prior to the formal evaluation. Specifically, we provided explicit anchor samples and reference videos with high and low consistency to calibrate participants' perception of Timbre Consistency and Spatial Alignment.

\textbf{Online Evaluation.} We conducted an online questionnaire-based evaluation with 53 participants, categorized into two groups: 23 film audio professionals (43.4\%) and 30 non-professionals (56.6\%). Participants evaluated stereo audio samples through a web-based interface. For baseline comparisons, we evaluated our FoleyDesigner against three state-of-the-art methods: See2Sound, Stable-Audio-Open, and SpatialSonic.

\subsection{Questionnaire Details}
\begin{figure}[h]
    \centering
    \includegraphics[width=0.9\linewidth]{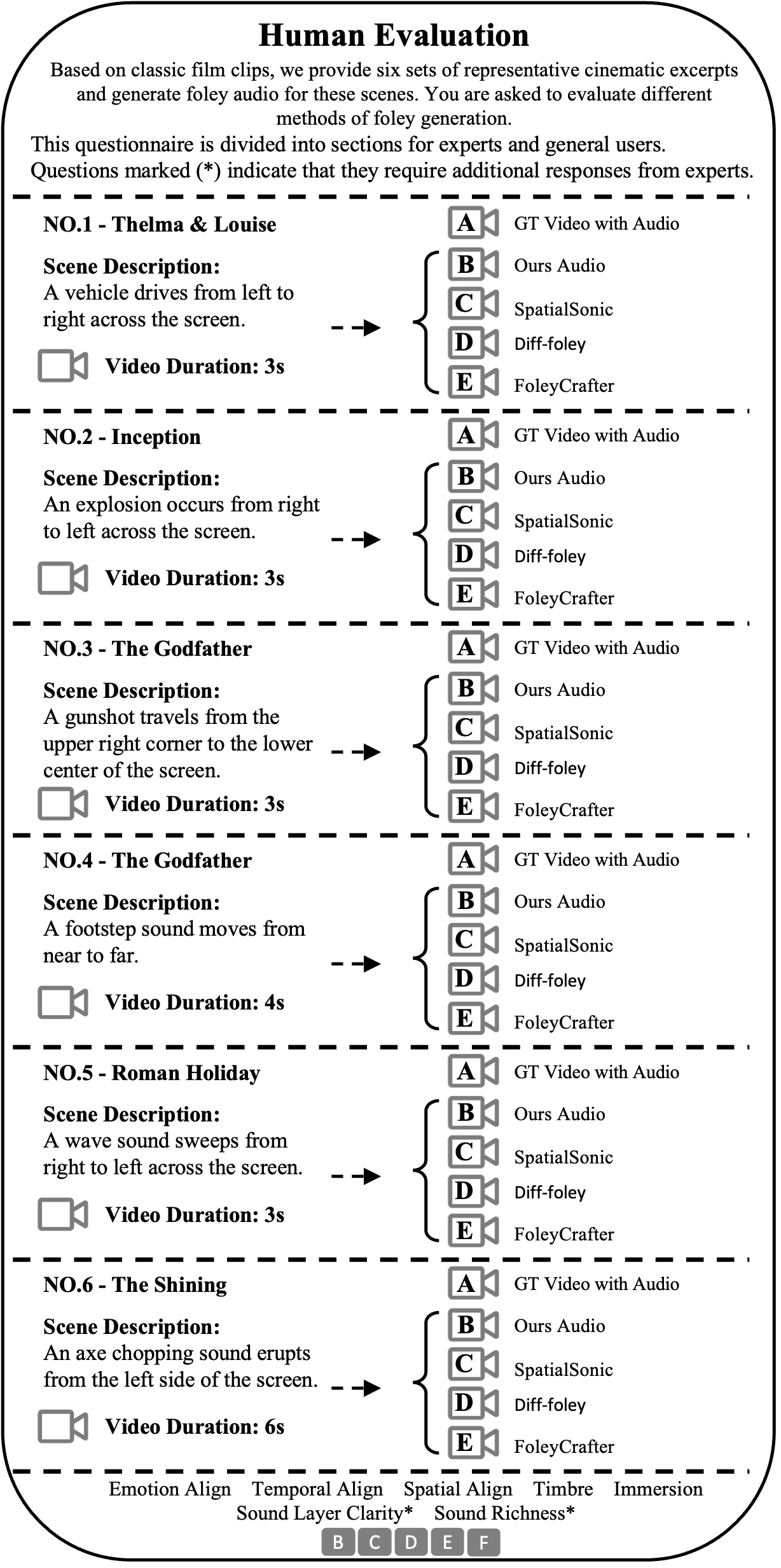}
\caption{\textbf{Questionnaire details}. This is the survey questionnaire we designed and used in the user study.}    \label{Questionnaire}
    \vspace{-4mm}
\end{figure}

Our online human evaluation was conducted through an questionnaire with 53 participants, categorized into two groups: 23 film audio professionals (43.4\%) and 30 non-professionals (56.6\%). The questionnaire was designed to evaluate four different audio generation methods across multiple criteria. Figure~\ref{Questionnaire} presents the questionnaire we designed for FoleyDesigner.

\textbf{For Non-Professional Participants:} The questionnaire consisted of five evaluation tasks using film clips. Participants were asked to select the best performing audio sample among four options based on the following criteria:
\begin{itemize}
\item\textbf{Timbral Matching}: Which audio has the highest compatibility between sound timbre and video content?
\item \textbf{Spatial Consistency}: Which audio demonstrates the highest consistency between sound spatial positioning and video?
\item \textbf{Temporal Alignment}: Which audio shows the highest consistency between sound timing and video?
\item \textbf{Emotional Coherence}: Which audio has the highest compatibility between sound emotion and video content?
\item\textbf{Immersiveness}: Which audio provides the strongest sense of immersion?
\end{itemize}

\textbf{For Professional Participants:} The questionnaire included two additional evaluation tasks using film clips, with extended criteria including:
\begin{itemize}
\item All criteria from the non-professional evaluation
\item \textbf{Audio Layering}: Which audio demonstrates the clearest hierarchy between primary and secondary sound layers?
\item \textbf{Detail Processing}: Which audio shows more refined detail processing?
\end{itemize}

Each participant watched the original film clips and then evaluated the four generated audio samples across all specified dimensions using a matrix single-choice format.

\subsection{Statistical Significance Analysis}
Since our user study is preference-based, we performed a Chi-Square Goodness-of-Fit Test to evaluate the statistical significance of the results. The analysis demonstrates a highly significant preference distribution ($p < 0.001$). Furthermore, a post-hoc Binomial Test confirms that our method significantly outperforms the second-best baseline with $p < 0.001$, indicating a robust consensus among raters.

\section{Case Study}

We conduct comprehensive qualitative analysis through two distinct case studies to evaluate temporal synchronization and spatial audio positioning capabilities across methods with different output channel configurations.

\subsection{Temporal Synchronization Analysis}

\begin{figure*}
    \centering
    \includegraphics[width=\linewidth]{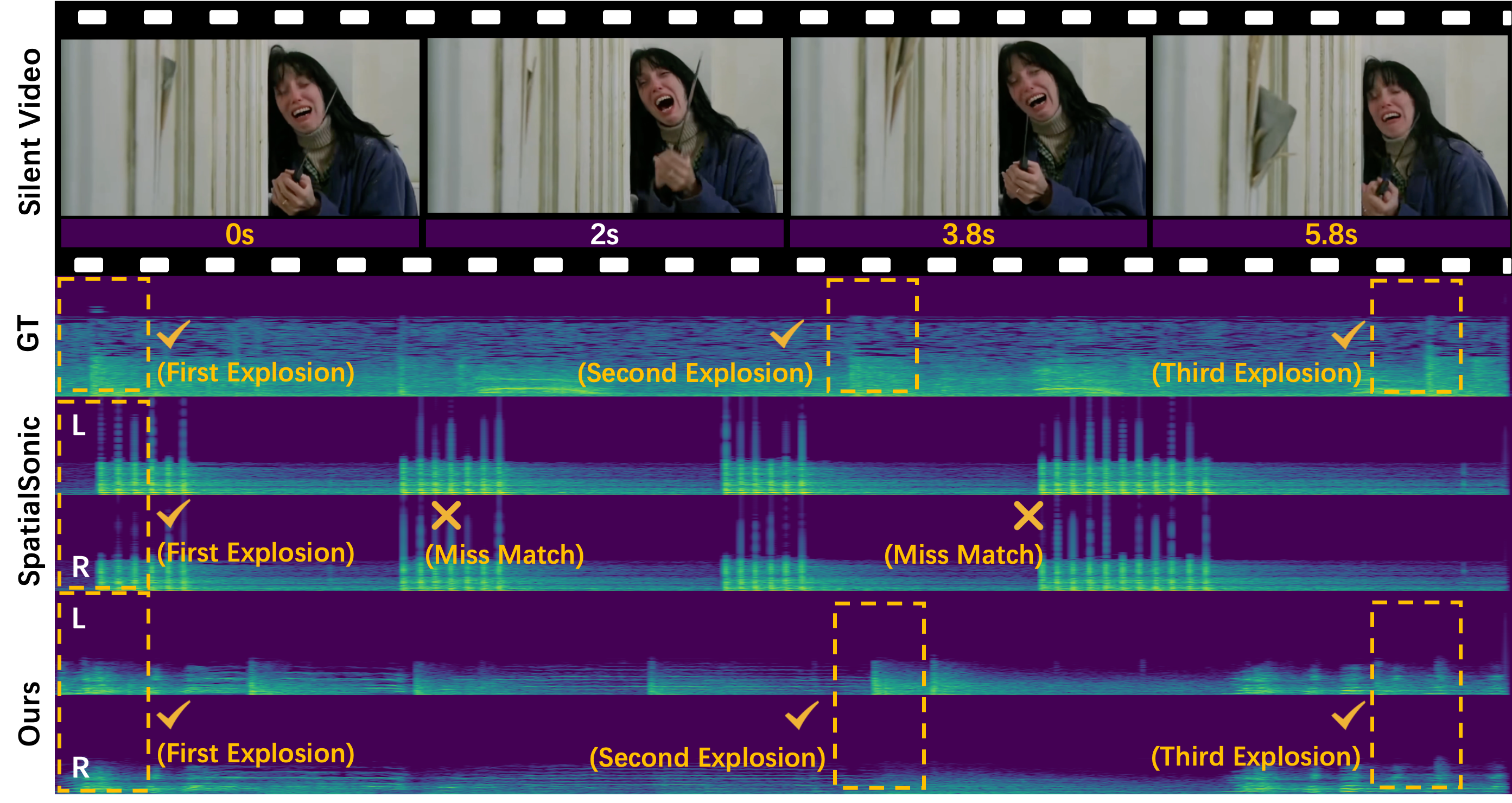}
    \caption{ \textbf{Temporal Analysis.} Each video frame corresponds to a temporal segment in the spectrogram below. Yellow checkmarks indicate successful audio-visual synchronization. Our method achieves consistent temporal alignment across key events, while baseline methods show varying degrees of synchronization failure regardless of their output channel configuration.}
    \label{fig:case_temporal}
\end{figure*}

Figure~\ref{fig:case_temporal} demonstrates the temporal alignment performance across different methods producing various audio formats. Each video frame in the input sequence corresponds to a specific temporal segment in the spectrogram visualization below. The yellow checkmarks indicate successful audio-visual synchronization points where the generated audio content accurately aligns with the visual events and their expected acoustic counterparts.

Our method, which outputs 5.1 Dolby surround audio, achieves great temporal consistency with checkmarks appearing at synchronization points throughout the sequence. This demonstrates that our approach successfully captures the timing of film sound events and generates corresponding audio that maintains precise temporal alignment.

As further evidenced by the quantitative results in Table~\ref{tab:baselines}, the baseline methods show varying temporal performance. SpatialSonic shows partial synchronization success, while the mono output methods (Diff-Foley and FoleyCrafter) demonstrate different temporal behaviors: Diff-Foley shows limited temporal coherence, missing several key synchronization opportunities, while FoleyCrafter achieves better alignment but still produces inconsistent temporal patterns compared to our multi-channel approach.

\subsection{Spatial Audio Positioning Analysis}

\begin{figure*}
    \centering
    \includegraphics[width=\linewidth]{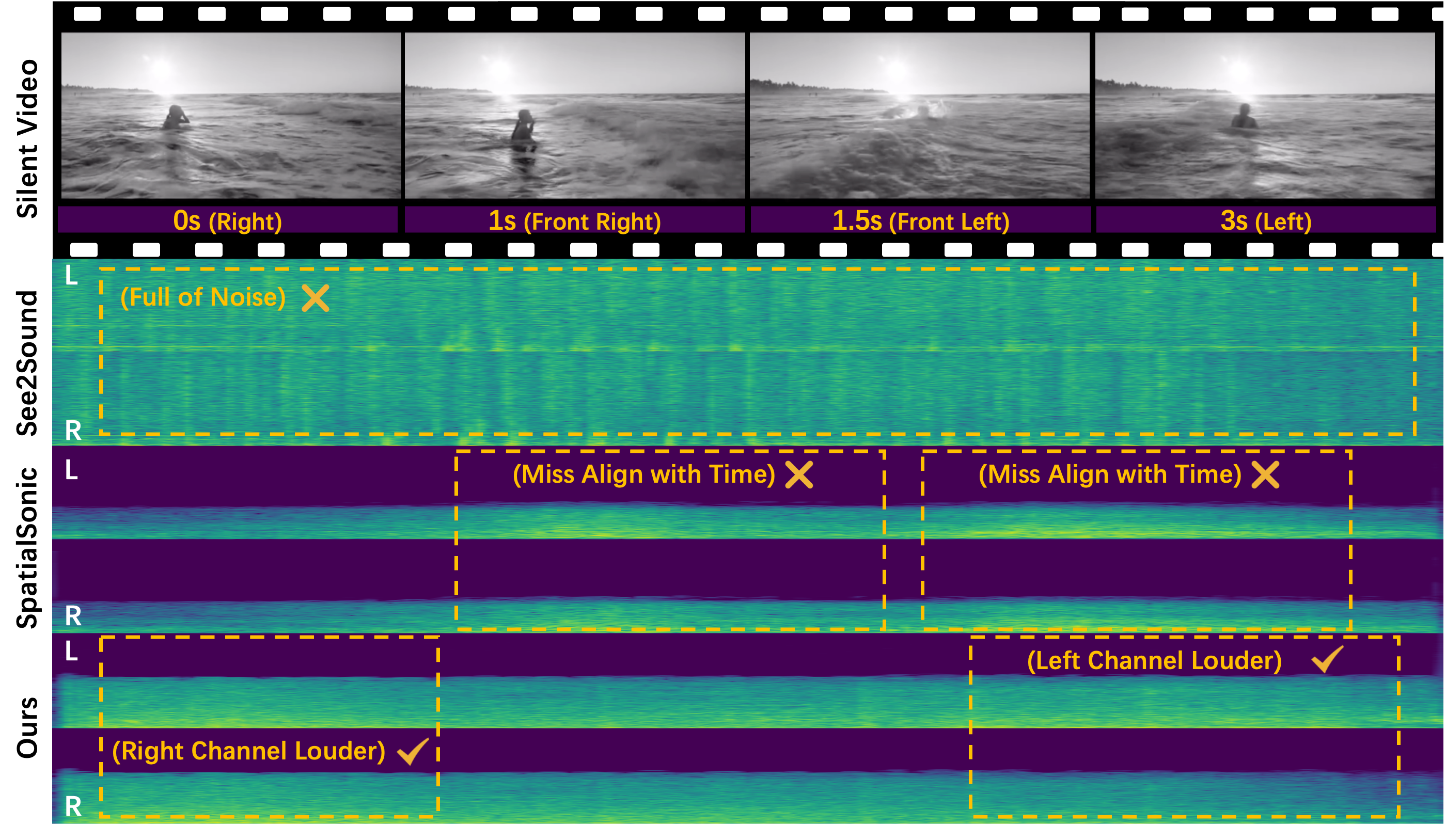}
    \caption{\textbf{Spatial Analysis.}  Our method demonstrates proper stereo separation with left channel (L) strengthening and right channel (R) weakening, while SpatialSonic (stereo output) shows limited spatial variation despite having two-channel capability.}
    \label{fig:case_spatial}
     \vspace{-4mm}
\end{figure*}

\begin{figure*}
    \centering
    \includegraphics[width=\linewidth]{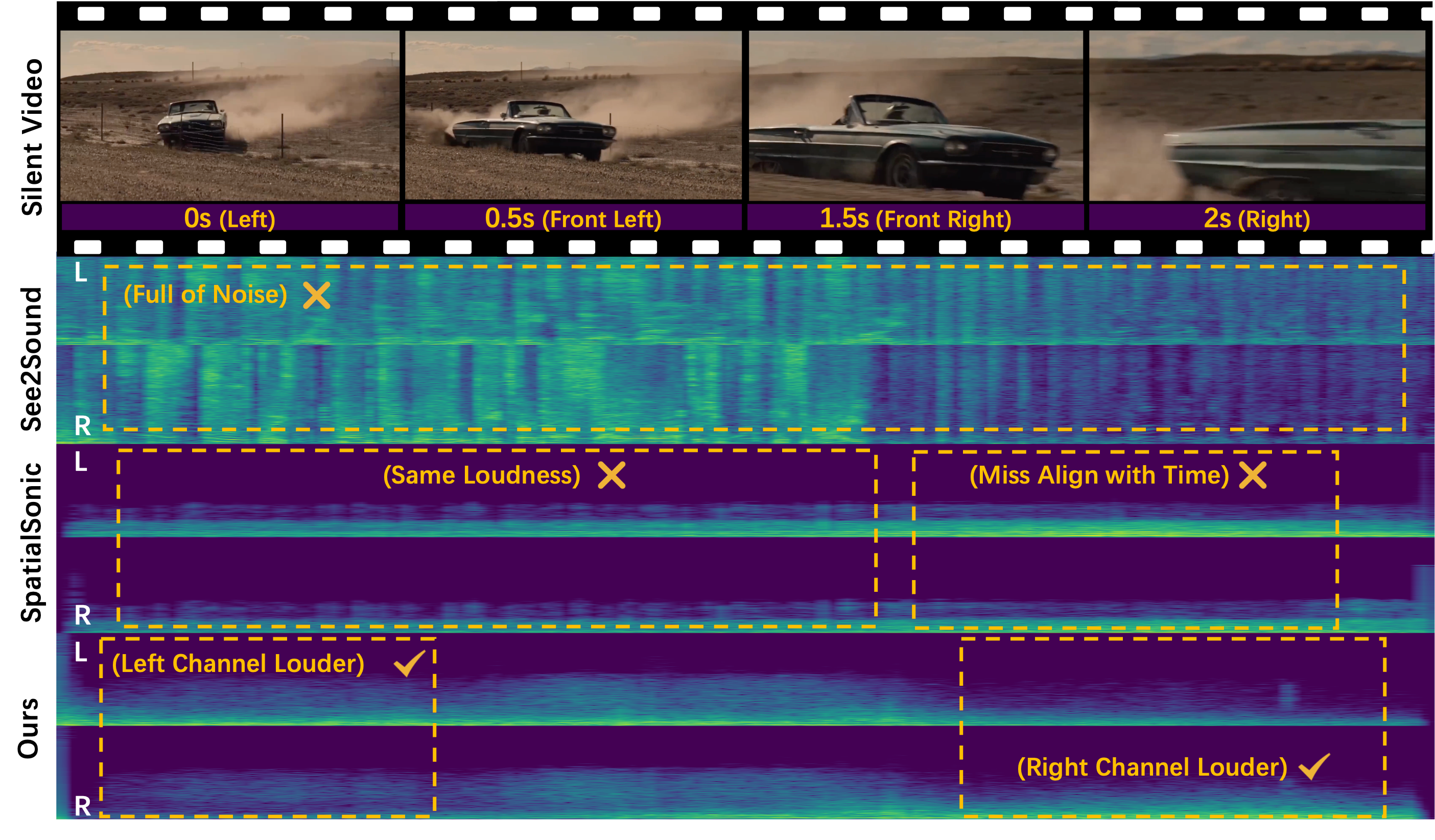}
    \caption{\textbf{Spatial Analysis.}  Our method demonstrates proper stereo separation with left channel (L) weakening and right channel (R) strengthening, while SpatialSonic (stereo output) shows limited spatial variation despite having two-channel capability.}
    \label{fig:case_spatial1}
     \vspace{-4mm}
\end{figure*}

Figures~\ref{fig:case_spatial} and~\ref{fig:case_spatial1} present spatial audio analysis demonstrating bidirectional spatial positioning capabilities. Figure~\ref{fig:case_spatial} shows a scene from Roman Holiday with ocean waves moving left-to-right, while Figure~\ref{fig:case_spatial1} demonstrates right-to-left movement.

Our method generates 5.1 Dolby surround audio with clear spatial positioning in both directions. In the left-to-right case, the left channel (L) gradually strengthens while the right channel (R) weakens. Conversely, in the right-to-left case, the left channel weakens while the right channel strengthens. This bidirectional channel variation accurately reflects the spatial movement of sound sources across scenes.

In contrast, SpatialSonic produces stereo output but exhibits limited spatial variation in both scenarios, with less pronounced channel differences that fail to capture the spatial movement. See2Sound generates audio  lacking spatial information.

These complementary examples demonstrate our method's robust spatial audio positioning across different movement directions.

\subsection{Discussion}

These case studies demonstrate our method's effectiveness in both temporal synchronization and spatial audio positioning. The temporal analysis shows consistent alignment with visual events, while the spatial analysis reveals appropriate channel separation that corresponds to visual movement. The comparison suggests that effective spatial audio generation requires not only multi-channel output capability but also proper modeling of spatial relationships in the audio generation process.

